\documentclass[journal]{IEEEtran}
\usepackage{times}
\usepackage{epsfig}
\usepackage{graphicx}
\usepackage{amsmath}
\usepackage{amssymb}
\usepackage{caption}
\usepackage{url}

\usepackage[pagebackref=true,breaklinks=true,letterpaper=true,colorlinks,bookmarks=false]{hyperref}
\usepackage{breakurl}
\usepackage{algorithm}
\usepackage{algorithmic}
\usepackage{subcaption}
\usepackage{multirow}
\usepackage{booktabs}
\usepackage{warpcol}
\usepackage{units}

\newcommand{\etal}{\mbox{\emph{et al.\ }}}
\newcommand{\ie}{\mbox{\emph{i.e.\ }}}
\newcommand{\eg}{\mbox{\emph{e.g.\ }}}
\newcommand{\wdist}{\ensuremath{\operatorname{\mathcal{W}}_2}}

\ifCLASSINFOpdf
\else
\fi
\begin{document}
	\onecolumn
	\pagestyle{empty}
\textcircled{c} 2015 IEEE. Personal use of this material is permitted. Permission from IEEE must be obtained for all other uses, in any current or future media, including reprinting/republishing this material for advertising or promotional purposes, creating new collective works, for resale or redistribution to servers or lists, or reuse of any copyrighted component of this work in other works.
	\clearpage
	\twocolumn
%
\title{Exploring Human Vision Driven Features for Pedestrian Detection}
%
%
%

\author{Shanshan~Zhang,
		Christian~Bauckhage,~\IEEEmembership{Member},
        Dominik A.~Klein,~\IEEEmembership{}		
        and~Armin B.~Cremers
\thanks{Manuscript received MONTH DAY, YEAR; revised MONTH DAY, YEAR.}
\thanks{S. Zhang and D. Klein are with the Department
of Computer Science III, University of Bonn, R\"omerstra\ss{}e 164, 53117 Bonn,
Germany (e-mail: zhangs@iai.uni-bonn.de, kleind@iai.uni-bonn.de).}
\thanks{C. Bauckhage is with Fraunhofer IAIS, Schloss Birlinghoven, 53757 Sankt Augustin, Germany (e-mail: christian.bauckhage@iais.fraunhofer.de).}
\thanks{A. B. Cremers is with Bonn-Aachen International Center for Information Technology (B-IT), Dahlmannstra\ss{}e 2, 53113 Bonn, Germany (e-mail: abc@iai.uni-bonn.de)}
}

%
%

\markboth{IEEE TRANSACTIONS ON CIRCUITS AND SYSTEMS FOR VIDEO TECHNOLOGY,~Vol.~X, No.~X, MONTH~YEAR}%
{Shell \MakeLowercase{\textit{et al.}}: Bare Demo of IEEEtran.cls for Journals}
%



\maketitle

\begin{abstract}
Motivated by the center-surround mechanism in the human visual attention system, we
propose to use average contrast maps for the
challenge of pedestrian detection in street scenes
due to the observation that pedestrians indeed exhibit discriminative contrast texture. Our main contributions are first to design a local, statistical multi-channel descriptor
in order to incorporate both color and gradient information. Second, we introduce a multi-direction and multi-scale contrast
scheme based on grid-cells in order to integrate
expressive local variations.
Contributing to the issue of selecting most discriminative features for 
assessing and classification, we perform extensive comparisons w.r.t. statistical descriptors, contrast measurements, and scale structures. This way, we obtain reasonable results under various configurations.
Empirical findings from applying our
optimized detector on the INRIA and Caltech pedestrian datasets show that our features
yield state-of-the-art performance in pedestrian detection.
\end{abstract}


%
\IEEEpeerreviewmaketitle

\section{Introduction}
\IEEEPARstart{T}{he} problem of pedestrian detection is attracting growing attention in the computer vision community as it has many practical applications in areas such as video surveillance or driving assistance systems. There is a quickly growing body of work on accurate and efficient detection of pedestrians in image or video data (see~\cite{Dollar2011} for a recent survey). Contributions have been made regarding problems such as feature extraction, classifier design, occlusion handling and the like. Although there were significant improvements over the last decade, one must acknowledge that the precision of state-of-the-art pedestrian detectors still lags behind human vision, which is capable of rapidly localizing pedestrians under various scales, poses, and occlusions even in low quality images. This motivates us to analyze how the human vision system processes incoming stimuli and to devise corresponding novel features for pedestrian detection. In this paper, we present experimental results which show that the use of biologically inspired mechanisms can indeed aid recognition.

\begin{figure}[t]
\centering
\includegraphics[width=0.43\textwidth]{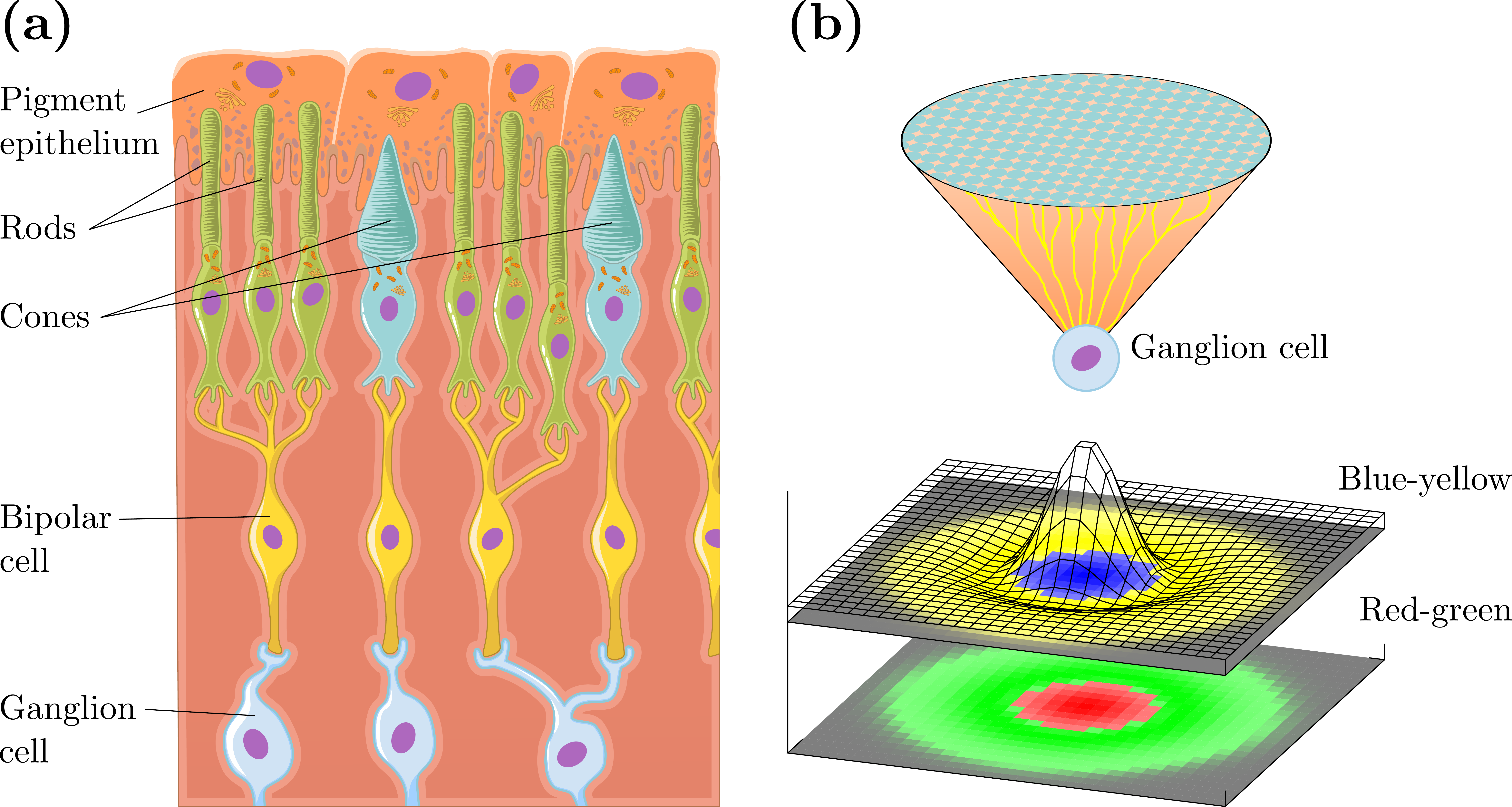}
\caption{Human retinal tissue: (a) Schematic cross-section (image adapted from~\cite{wiki_RodsCones}); (b) Spatial wiring and DoG weighting of retinal ganglion cells.}
\label{fig:retina_ganglion}
\end{figure}

In the human visual system, processing of information begins in the retinal tissue immediately after photoreceptive cells (rods for lightness resp. cones for colors) have transformed incident light into electric signals (cf. Fig.~~\ref{fig:retina_ganglion}). In a first layer of bipolar cells, electrical membrane potentials are locally aggregated. Grouped bipolar cells report to different types of ganglion cells, which convert analog potentials into electric pulse rates. At the transitional synapses between photoreceptive and bipolar cells, but also from bipolar to ganglion cells, there is a lateral wiring of so called horizontal respectively amacrine cells modulating the signals to enhance contrasts in a center-surround fashion. It was found that the output of certain ganglion cells agrees with simple difference of Gaussian (DoG) filter responses~\cite{Rodieck1965} or more complex oriented Gabor filter results~\cite{JonesPalmer1987}. A more in-depth survey on retinal cell types and their wiring can, for instance, be found in \cite{Lee2010}.

The center-surround mechanism is also found in later processing stages in the brain where it guides human \textit{attention} and thus affects how people recognize objects of interest. This psychophysical theory has been widely used in computational approaches to generate saliency maps of the environment~\cite{Frintrop2010}. However, while attention is about bottom-up processing, model-free analysis of signals from the environment, visual search for specific entities requires top-down saliency which tunes the scoring of basic features to the expected appearance.

In this article, we propose to emulate center-surround contrast features motivated by the human visual system and to tune them towards characterizations of the appearance of pedestrians.
Our previous findings about human vision driven features were
published in \cite{Zhang2014ICPR}; in this paper, we explore the configurations of feature design and achieve better performance.
Our contributions are summarized as follows:

\textbf{Statistical multi-channel cell descriptors:} We collect multi-channel information for each cell area, \ie local image patch, not only regarding lightness and colors, but also w.r.t. gradients which complement each other in
recognizing broad variations of clothing or articulations of the human body. In order to summarize the underlying, unknown distribution of each cell's channel values, we propose two kinds of distributions: (1) a continuous Gaussian distribution which models maximum entropy given a known mean and variance~\cite{CoverThomas2006}; (2) a bilinear interpolated histogram which is a representation of frequencies observed over discrete intervals (bins).

\textbf{Multi-direction and -scale contrast vectors:} Aiming at incorporating more specific information between central and surrounding cells, we treat adjacent image regions in different directions individually rather than as a single surrounding region and thus obtain multi-direction contrast descriptors; we compute statistical features at different cell-sizes so as to build a contrast pyramid which accords with the general architecture of most visual saliency systems.

\textbf{Extensive evaluations on various configurations:} In order to determine the strongest feature scheme for pedestrian detection, we implement various contrast measurements for both distributions and at different scale structures. In extensive evaluations on the INRIA dataset, we find that the
advisable scheme is to use a Gaussian-\wdist~combination and a 4-6-8-10 scale structure.

Our presentation proceeds as follows: Section~\ref{sec:sec_related_work} provides an overview of related work; Section~\ref{sec:sec_overview} presents details as to our feature extraction mechanism. Two key components of this mechanism, namely statistical descriptors and contrast
measures are explained in Section~\ref{sec:sec_descriptor} and Section~\ref{sec:sec_measurement}, respectively. Our classification procedure is presented in Section~\ref{sec:sec_classification}, followed by a discussion of thorough and extensive experiments in Section~\ref{sec:sec_experiments} where we compare different feature schemes and state-of-the-art detectors on standard benchmarks. Finally, we conclude and propose several directions for future work in Section~\ref{sec:sec_conclusion}.

\begin{figure}
\centering
\begin{subfigure}[b]{0.4\textwidth}
	\centering
	\includegraphics[width=\textwidth]{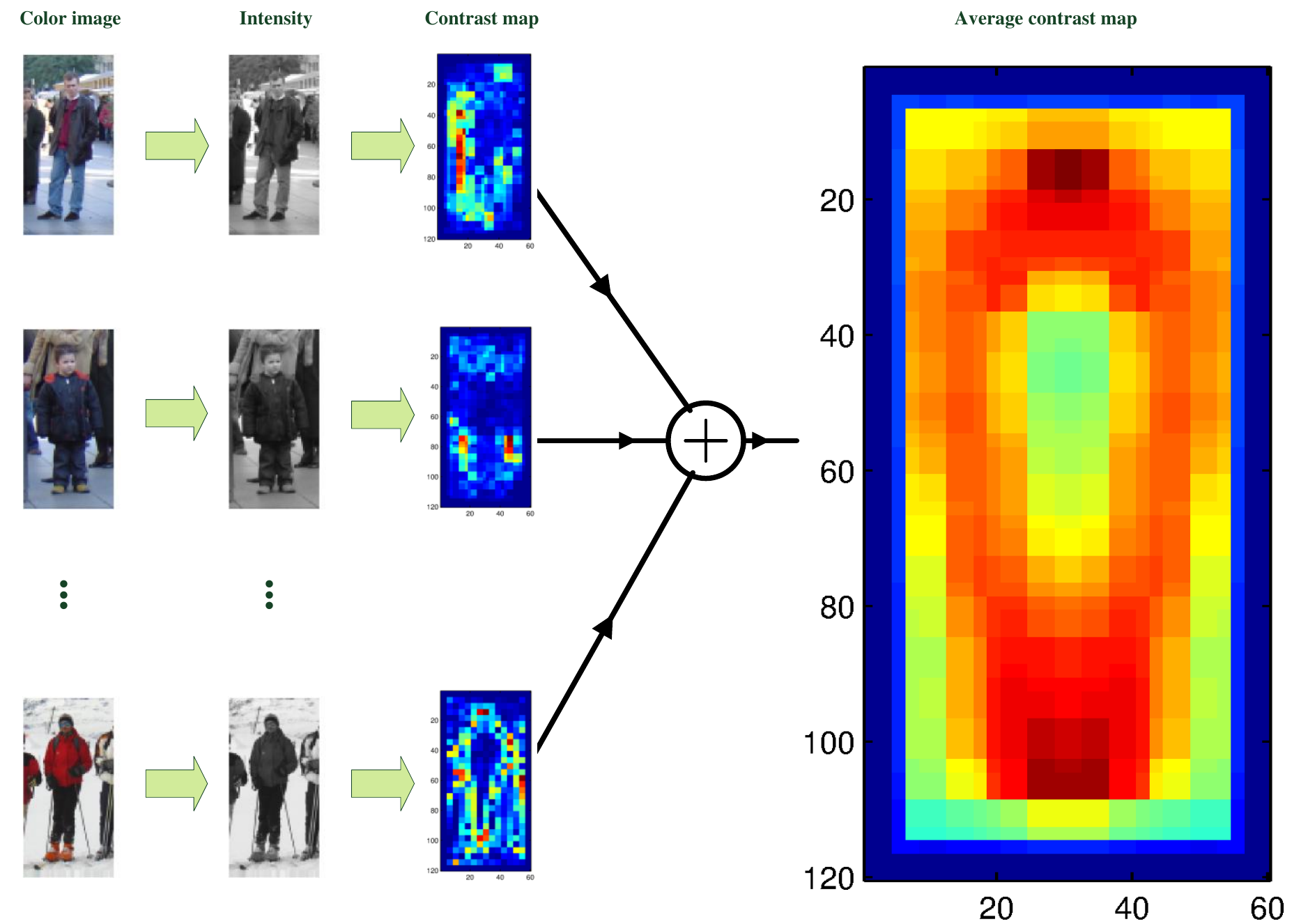}
	\caption{Average contrast map for pedestrians}
	\label{fig:fig_avg_contrast_pos}
\end{subfigure}
\hfill
\begin{subfigure}[b]{0.4\textwidth}
	\centering
	\includegraphics[width=\textwidth]{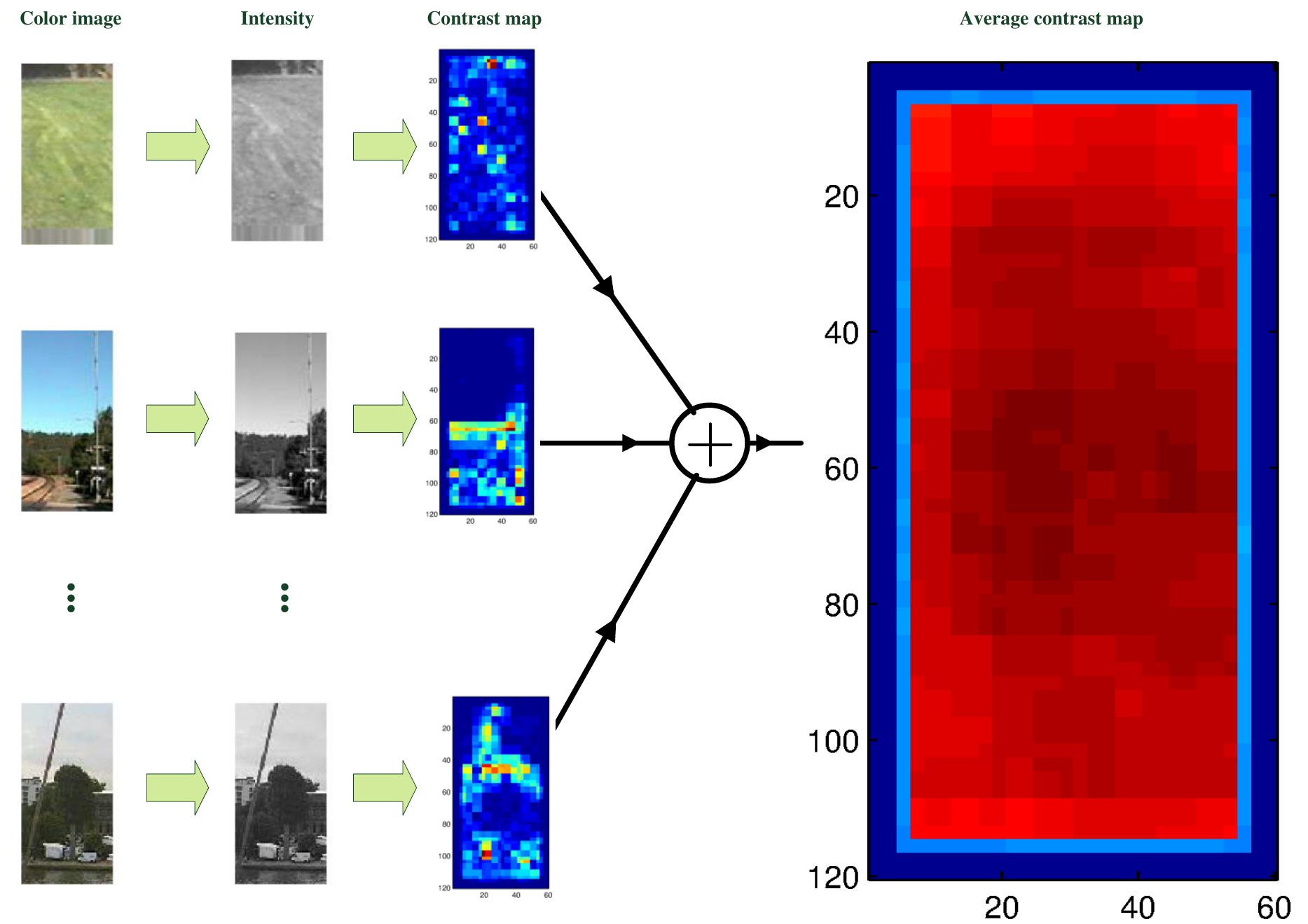}
	\caption{Average contrast map for non-pedestrian examples}
	\label{fig:fig_avg_contrast_neg}
\end{subfigure}
\caption{Heat maps of average center-surround contrasts generated from positive and negative samples of the INRIA pedestrian dataset. Warmer colors indicate higher contrast values.}
\label{fig:fig_avg_contrast}
\end{figure}

\begin{figure*}[t]
\centering
\centerline{\includegraphics[width=0.8\textwidth]{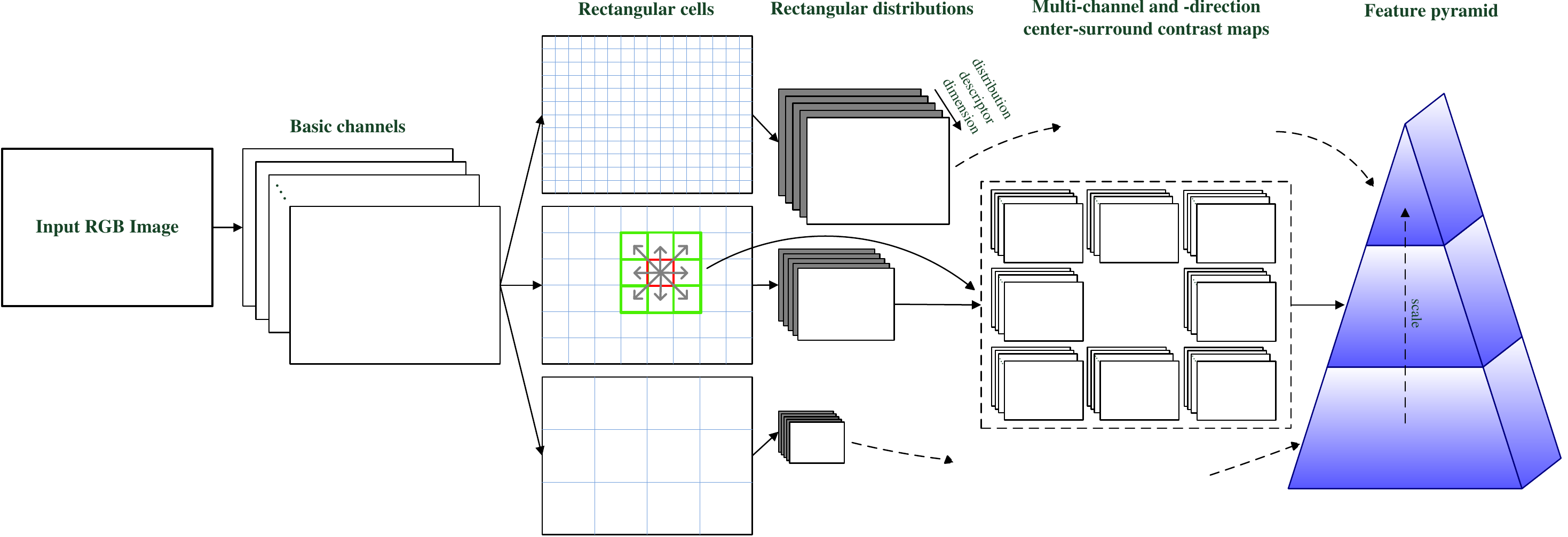}}
\caption{Flow chart of our feature extraction mechanism. Here, we consider a three-scale structure as an example but different scale structures can be used as well.}
\label{fig:fig_feature_extraction}
\end{figure*}

\section{Related work}
\label{sec:sec_related_work}
Since our focus in this paper is on emulating the center-surround mechanism in human vision in order to design new contrast features for pedestrian detection, the following literature review mainly considers difference based features for pedestrians and center-surround contrast
measures used by computational visual attention approaches.

\subsection{Features for pedestrian detection}
Most features for pedestrian detection interpret local or global pixel differences in various forms. This is because pixel differences represent texture information which are often characteristic for classes of objects and thus allow for robust classification.

Gradients 
(vectors of directed derivatives) are popular features as they describe differences w.r.t. intensity or colors between neighboring pixels and allow for characterizing these in terms of magnitudes and orientations. The arguably most popular kind of feature \textit{Histograms of Oriented Gradients} (HOGs)~\cite{Dalal2005} for pedestrian detection is indeed built on gradient statistics. HOG features brought about significant improvements and therefore establish an important baseline. Several researchers have extended this feature pool and added further features. For example, Liu \etal \cite{Liu2009} introduced the idea of a granularity space, \ie a family of descriptors ranging from edgelets to HOGs.

\textit{Local Binary Pattern} (LBP) features~\cite{Ojala1996} are another kind of pixel-wise difference based features which
express relative intensity relationships between neighboring pixels by binary codes. Wang \etal ~\cite{Wang2009} combined LBP features with HOG features in order to better cope with occlusions; Ma \etal ~\cite{Ma2013} proposed a set of edge orientation histogram (EOH) and oriented LBP based features to describe cell-level and block-level structure information.

\textit{Haar-like} features~\cite{Viola2001}, on the other hand, are considered as patch-wise local differences as they compute sums of intensity values over rectangular image regions. Zhang \etal~\cite{Zhang2014CVPR} designed Haar-like templates tailored to up-right human body and achieved significant improvement.

\textit{Color Self Similarity} (CSS) features proposed by Walk \etal~\cite{Walk2010} describe global differences between pairs of image cells in terms of color histograms. Significant improvements were achieved by combining HOG features and CSS features, since they allow for representing uniform textures found in people's clothing.

Although extensive efforts have been made to interpret local difference in various ways, the performance of all the above features is still far behind humans' capabilities. We therefore argue that it is worthwhile to look into how the human brain processes visual inputs and next to
mimic corresponding mechanisms in order to design more representative features.

To our knowledge, the first attempt of designing human vision inspired features dedicated to pedestrian detection can be found in \cite{Montabone2010}.
Although our motivation in this article is a similar one, our features considerably differ from the ones in \cite{Montabone2010}.
The following clear distinctions can be drawn: 1)~we consider local difference between central and surrounding square regions rather than between pixels; 2)~we compute the center-surround contrasts in multiple channels (not only on colors but also on gradient information); 3)~we do not use image channel values directly but describe their distributions using statistical entities; 4)~we do not treat neighboring regions as a whole but individually and thus incorporate more detailed information regarding local difference.

\subsection{Center-surround contrast measures}
Most computational approaches to visual attention determine center-surround contrasts by DoG-filters or approximations of these~\cite{Itti1998}. Recently, several researchers represented the central and surrounding areas in terms of feature distributions so as to capture more information about the areas. These distributions were either discrete, \eg in form of  histograms~\cite{Klein2011}, or continuous, \eg
fitted to a normal distribution~\cite{Klein2012}, and various distance measures can be
applied between central and surrounding distributions to quantify local contrast.

However, we notice that the above
strategies only achieve reasonable results for rather
conspicuous scenarios, \eg a big red flower standing out in surrounding green leaves. In fact, the background in our case is much more complex and the previous
contrast models are not guaranteed to perform well. Consequently, we train and evaluate
specialized contrast schemes in this paper and aim to find out the optimal
configuration for our applications.

\section{Overview on feature extraction}
\label{sec:sec_overview}

This section introduces the feature extraction procedure we consider in this paper.
First of all, we demonstrate that center-surround contrasts are discriminative for pedestrians. We collect $2,416$ positive samples (consisting of pedestrians) and $5,000$ negative samples (no pedestrians included) from the INRIA pedestrian dataset \cite{Dalal2005}. All the sample color images are converted to gray images since
considering only
intensity
is sufficient to showcase the possible performance gain. The contrast map of each image is computed on two scales of $4\times 4$ and $6\times 6$ pixels. The contrast value, represented by the difference between central and surrounding cell regions w.r.t. mean value, is added to the central region. Finally, two average contrast maps are generated for the pedestrian class and non-pedestrian class, respectively. In Fig.~\ref{fig:fig_avg_contrast}, we see that the average contrast map for pedestrians indeed resembles a human body while the average contrast map for non-pedestrians shows no 
distinct pattern.

Based on the above observation, we design our center-surround contrast features for pedestrian detection.
An illustration of our feature extraction procedure is shown in Fig.~\ref{fig:fig_feature_extraction}. First, we compute multiple channels (\eg color and gradient information) for each pixel in an image; second, we divide each channel map into square cells of a fixed size and describe each cell using statistical distributions; third, we compute the differences between each cell and its eight nearest neighboring cells so as to obtain a multi-direction contrast vector; finally, we repeat the second and third step along each channel with different cell sizes and thus obtain a multi-channel, multi-direction, and multi-scale contrast pyramid for the whole image.

\subsection{Center-surround contrasts}
The
core part of our feature extraction is how to
determine the difference between two rectangular regions. To address this problem, we first choose an appropriate distribution for each region, as discussed (see Section~\ref{sec:sec_descriptor}) and then consider corresponding contrast measures to numerically describe the difference between two given distributions (see Section~\ref{sec:sec_measurement}). In order to determine the strongest center-surround contrast features for pedestrian detection, we conduct extensive experiments and comprehensive comparisons on various combinations of distributions and contrast measures. Experimental results under different schemes are presented in Section~\ref{sec:sec_experiments}.

\subsection{Channels}
To consider multiple feature channels in our scheme is motivated by the success of Doll\'ar's detector [ChnFtrs]~\cite{Dollar2009a}, which has been established a strong baseline due to its accuracy and efficiency. In [ChnFtrs], multiple channel maps, in terms of colors and gradients, are computed for each input image, and the final feature values consist of local sums at different spatial locations and over all channel maps. These local sums are efficient to compute by employing integral images. They are less sensitive to noise than the 
individual channel values.

\begin{figure}
\centering
\centerline{\includegraphics[width=0.3\textwidth]{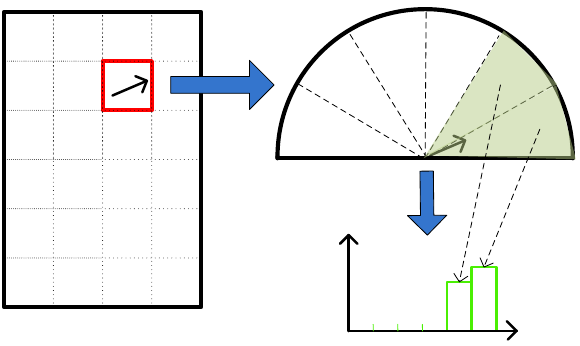}}
\caption{Illustration of a histogram of oriented gradients obtained for a single pixel. }
\label{fig:fig_orientation_hist}
\end{figure}

Similar to [ChnFtrs], we also consider a total of 10 different
channels: 3 channels for LUV colors, 1 channel for gradient magnitude information, and 6 channels for histograms of oriented gradients.
Note that all the above channels are computed pixel by pixel. Histograms of oriented gradients are usually computed for a group of pixels inside an image region, but we compute them
pixelwise which is to say we simply quantize the gradient magnitudes into orientation bins. For each pixel, two neighboring bins are 
affected as we employ bilinear interpolation w.r.t. orientation bins, see Fig.~\ref{fig:fig_orientation_hist} for an illustration.

Prior to channel computation, input images are smoothed with a binomial filter~\cite{Haddad1971} of radius 1, \ie~$\sigma \approx 0.87$, in order to remove noise. Note that we explicitly do not smooth channel data as we observed this to lead to decreased performance.

\subsection{Center-surround neighborhood patterns}
Here, we present details on our design of center-surround cell pairs. Four patterns are proposed in this paper and explained in the following.

\textit{$C_1S_8$ pattern:} For each cell, its eight nearest neighboring cells are considered as surrounding cells, denoted as $[C^s_1, C^s_2,...,C^s_8]$. The eight surrounding cells can be treated either as a whole or separately. From our experiments, we find a significantly better performance if they are treated individually (cf. Fig.~\ref{fig:fig_cspatterns}), since difference information in eight directions are integrated respectively. Thus, we use this \textit{$C_1S_8$} pattern in our experiments to build a multi-direction contrast vector for each cell along
every channel.

\begin{figure}[]
	\centering
	\begin{subfigure}[t]{0.17\textwidth}	
		\centering
		{\includegraphics[width=\textwidth]{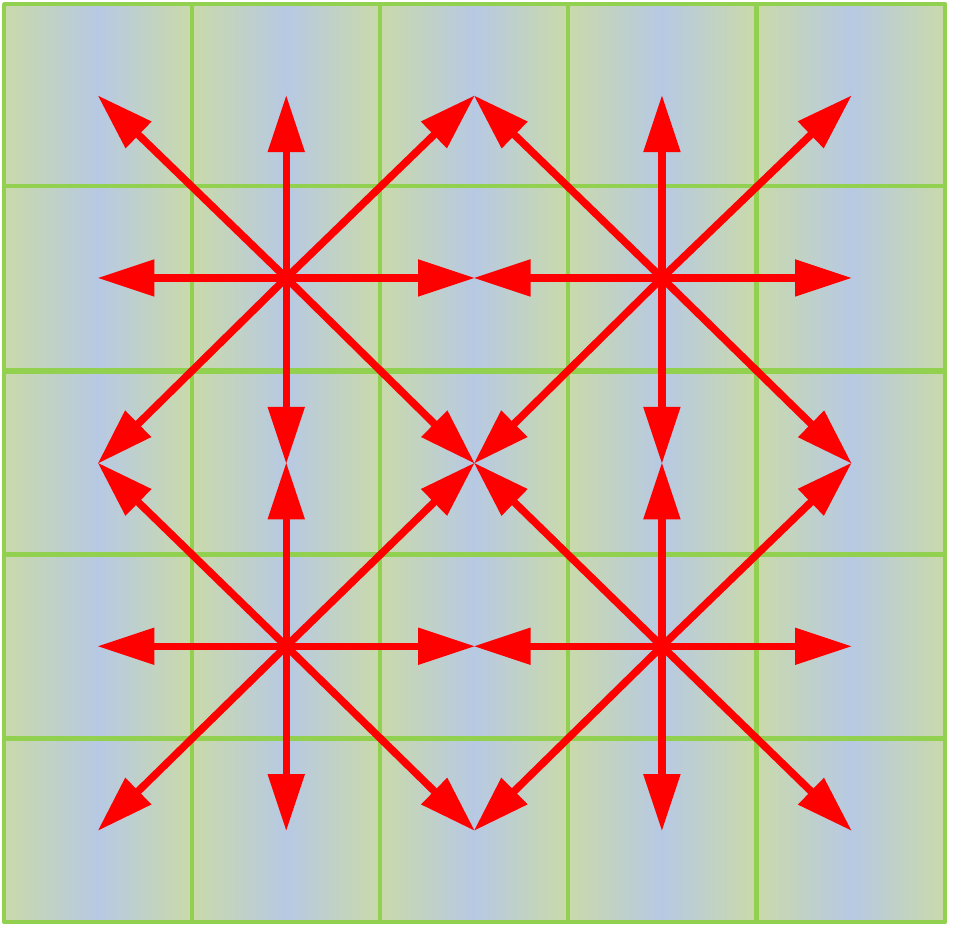}}
		\caption{}
		\label{fig:fig_neighborhood_maps}
	\end{subfigure}
	\hspace{10pt}
	\begin{subfigure}[t]{0.17\textwidth}
		\centering
		{\includegraphics[width=\textwidth]{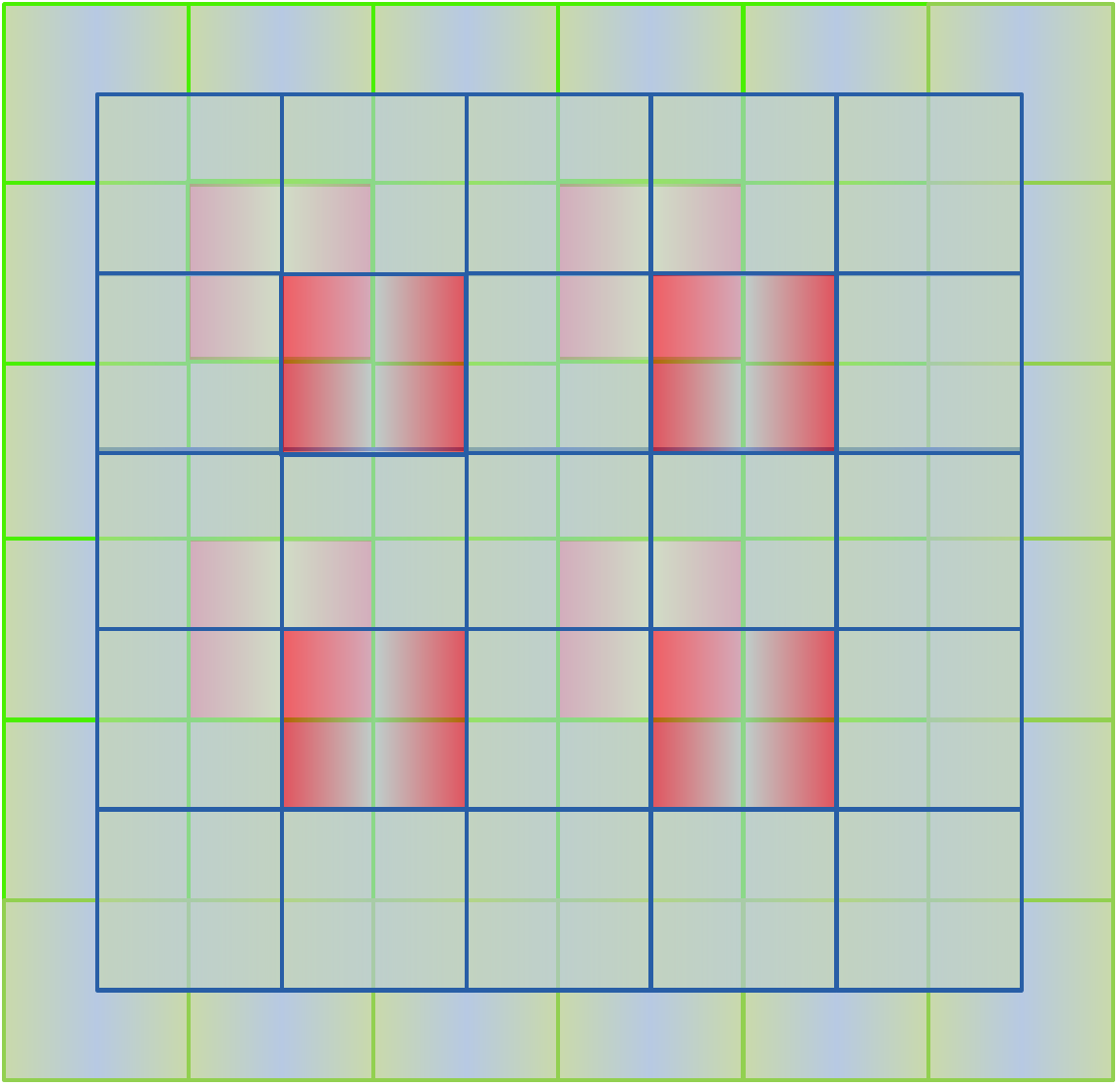}}
		\caption{}
		\label{fig:fig_shift}
	\end{subfigure}
	\caption{Illustration of two neighborhood patterns. (a) Sparse pattern. Each red arrow points from the central cell to one of its neighboring cells. (b) Shift pattern. Two layers of cells are denoted with green and blue grid lines and the red cells denote central cells with eight nearest neighboring cells.}
	\label{fig:fig_neighbor_patterns}
\end{figure}

\textit{Sparse pattern:} Significant redundancy will emerge if we consider eight nearest neighboring cells for each cell, because each adjacent pair of cells is
incorporated twice. To avoid this redundancy, we use a cell step of 2 cells along both horizontal and vertical directions, resulting in a sparse
neighborhood map as shown in Fig.~\ref{fig:fig_neighborhood_maps}.

\textit{Shift pattern:} According to the Nyquist-Shannon sampling theorem \cite{Shannon1949}, we propose a shift mechanism where we define two cell layers and iterate the \textit{$C_1S_8$} center-surround pattern on each respectively. For the first layer, we start from the left top pixel and divide the whole image into square cells; for the second layer, the starting point is shifted with a
step of $0.5$ times the cell size along both horizontal and vertical directions and we then divide the image patch from the new starting point into square cells with the same cell size. An illustration of the shift mechanism is shown in Fig.~\ref{fig:fig_shift}.


\textit{Multi-scale pattern:} Finally, in order to describe contrasts at different scales, we use different cell sizes to build a contrast pyramid which is in accordance with the general architecture of most computational visual attention systems.

\section{Statistical cell descriptors}
\label{sec:sec_descriptor}
In order to
assess the
underlying distribution inside each cell, we
estimate both continuous and discrete statistical
approximations: (1) a Gaussian distribution, which is the type of continuous distribution with maximum entropy given a known mean and variance; (2) a bilinear interpolated histogram, which is a representation of frequencies, determined for discrete intervals (bins).

In our following discussion, we assume that we have
measured values
of channel $i$ for the whole input image. We denote this data as channel image $P^i$ and consider a specific cell $c$ with its channel vector $P_c^{i} = [v_1^{i},v_2^{i},...,v_p^{i}]$.

\subsection{Gaussian distributions}
The true distribution of channel values for local image patches is unknown, but is modeled as Gaussian type in this section. This assumption is made not only because normality makes further estimations convenient to solve, but also due to its popularity in classic low-level vision models, for example, in \cite{Horn1980}.


For numerical description, we apply maximum likelihood (ML) estimation of the parameters and obtain mean and variance values as
\begin{equation}
\hat{\mu}_c^i = \frac{1}{p}\sum\limits_{k=1}^{p} v_k^{i} = \overline{P_c^i},
\label{eq:eq_mu}
\end{equation}
and
\begin{equation}
\hat{\Sigma}_c^i = \frac{1}{p}\sum\limits_{k=1}^{p} (v_k^{i}-\overline{P_c^i})^2 = \overline{(P_c^i)^2} - \overline{P_c^i}^2.
\label{eq:eq_sigma}
\end{equation}

Now the estimation is narrowed down to computing two local averages: $\overline{P_c^i}$ and $\overline{(P_c^i)^2}$ according to Eq.~\ref{eq:eq_mu} and Eq.~\ref{eq:eq_sigma}. For efficiency, we employ two integral images
for each channel: one for the original channel image $P^i$ and the other for the squared channel image $(P^i)^2$ and thus avoid extensive summations per individual cell.

Once the parameters of the Gaussian have been determined, we obtain a descriptor for cell $c$
for channel $i$:
\begin{equation}
D^i(c) = [\mu_c^i, \Sigma_c^i].
\label{eq:eq_gaussian_descriptor}
\end{equation}

\subsection{Histograms}
Histograms are a reasonable discrete representation of distributions without any prior assumption of the underlying statistics. They count the observed frequencies of data appearing in discrete intervals. The advantage of using a histogram is furthermore that it tolerates noise and minor intra-class differences and that its degree of tolerance can be adjusted by choosing appropriate numbers of bins. Generally, using a smaller number of bins results in a coarser description of the original data, and vice versa.

It is computationally expensive to naively compute histograms for all cells and all sizes so that we employ integral histograms \cite{Porikli2005}. An integral histogram can be considered as a stack of integral images each counting the sums of values to the top and left from a pixel that fall into a certain histogram bin.

To eliminate bias, we implement bilinear interpolation for histograms, \ie each value contributes into two nearest bins with a weight relating to distance between the given value and the bin center. Also, normalization is rather important for histograms, since it eliminates the effect of data magnitudes. In this paper, we normalize each local histogram for each cell
and channel so that it sums up to 1. In the end, given $b$ bins, we obtain a histogram $H^i_c$ as a descriptor vector for channel vector $P^i_c$:
\begin{equation}
H^i_c = [h^i_c(1),h^i_c(2),...,h^i_c(b)], \sum\limits_{k=1}^{b}h^i_c(k)=1.
\label{eq:eq_histogram_descriptor}
\end{equation}

\section{Contrast measurements}
\label{sec:sec_measurement}
Aiming for the strongest center-surround contrast features, we introduce multiple contrast measurements for each distribution descriptor to make a comprehensive comparison in this section. Combining a distribution descriptor and a corresponding measurement forms a specific scheme for feature extraction. We note that the cell descriptors introduced above are statistical distributions whose comparison requires care. Although the Euclidean distance is often used in practice, it is not truly faithful to the nature of this kind of data. In particular, when comparing distributions or histograms, we are dealing with compositional data \cite{Aitchison1986-TSA}. This is to say that, for a normalized histogram $H = [h(1), \ldots, h(b)]$ of $b$ bins, there are only $b-1$ degrees of freedom, since the value of an arbitrary bin $h(i)$ is determined by $1 - \sum_{k \neq i}^b h(k)$. It is therefore impossible to perturb one bin of a histogram without affecting the others. This has implications for similarity measurements that are not accounted for by the Euclidean metric. However, there are several distance- or similarity measures that cope with these characteristics and we consider their use in our context. To summarize, we explore six different metrics in this paper: Gaussian-\wdist~distance, Gaussian-$L^2$ distance, Gaussian gradient matrices, histogram Kullback-Leibler divergences, histogram Hellinger distances, and histogram intersections.

In the following, we denote the
channel distributions for a central and a surrounding cell as $P^i_c$ and $P^i_s$, respectively. The contrast vector $\overrightarrow{cst}(P^i_c,P^i_s)$ is computed using different measures.

\subsection{Gaussian distributions}
We introduce three different contrast measures to compute the difference between two cells'
channel distributions, each represented by the Gaussian descriptor in Eq.~\ref{eq:eq_gaussian_descriptor}. We compare the results of those three measures in Section~\ref{sec:sec_experiments}.

\subsubsection{\textbf{\wdist~distance}}
The \wdist~distance (2nd Wasserstein distance) was first introduced as a measure for center-surround contrast by Klein \etal~\cite{Klein2012} and achieved reasonable results for saliency detection. Its definition in our case can be written as:
\begin{equation}
\begin{split}
\wdist(P_{c}^{i},P_{s}^{i}) =
\bigg[\inf_{\gamma\in\Gamma(P_{c}^{i},P_{s}^{i})}\int_{\mathbb R \times \mathbb R}|x-y|^{2}\operatorname{d}\gamma(x,y)\bigg]^{\frac{1}{2}},
\end{split}
\label{eq:eq_wasserstein}
\end{equation}
where $\Gamma(P_{c}^{i},P_{s}^{i})$ denotes the set of all couplings of $P_{c}^{i}$ and $P_{s}^{i}$.

It would be intractable to compute the integral in Eq.~\ref{eq:eq_wasserstein} in case of arbitrary distributions. However, for the Gaussian distribution, it can be solved analytically \cite{Givens1984}. The contrast vector between one central cell distribution $P^i_c\sim N(\mu_c^i,\Sigma_c^i)$ and its neighboring cell distribution $P^i_s\sim N(\mu_s^i,\Sigma_s^i)$
of channel $i$ indeed amounts to:
\begin{equation}
\begin{split}
\wdist(P_{c}^{i},P_{s}^{i}) = \bigg[||\mu_c^i - \mu_s^i||_2^2 +
 \Sigma_c^i + \Sigma_s^i
-2\sqrt{\Sigma_c^i \Sigma_s^i}\bigg]^{\frac{1}{2}}.
\end{split}
\label{eq:eq_wasserstein_gaussian}
\end{equation}

We note that Wasserstein distances are natural metrics for the comparison of two probability distributions where one distribution is derived from the other one through small, non-uniform perturbations; in the computer vision literature, the discretized Wasserstein distance is also referred to as the Earth Mover's distance \cite{Rubner1998-AMF}.

\subsubsection{\textbf{$L^2$ distance}} If we treat the two-dimensional descriptors for the central and surrounding cells as two 2D points, then the $L^2$ distance between $(\mu_c^i,\Sigma_c^i)$ and $(\mu_s^i,\Sigma_s^i)$ amounts to:
\begin{equation}
D_{L^2}(P_{c}^{i},P_{s}^{i}) = \sqrt{(\mu_c^i - \mu_s^i)^2+(\Sigma_c^i - \Sigma_s^i)^2}.
\label{eq:eq_L2}
\end{equation}

\subsubsection{\textbf{Signed gradient matrix (SGrd)}}
For each center-surround cell pair, we compute the signed gradient matrix for the mean and variance vector $[\mu^i,\Sigma^i]$, resulting in a contrast vector. The contrast vector between one central cell distribution $P^i_c\sim N(\mu_c^i,\Sigma_c^i)$ and its neighboring cell distribution $P^i_s\sim N(\mu_s^i,\Sigma_s^i)$
of channel $i$ can then be expressed as follows:
\begin{equation}
\begin{split}
\overrightarrow{SGrd}(P_{c}^{i},P_{s}^{i}) = \bigg[\mu_c^i - \mu_s^i, \Sigma_c^i - \Sigma_s^i \bigg].
\end{split}
\label{eq:eq_grd_matrix}
\end{equation}

In the feature space, the contrast vector in Eq. \ref{eq:eq_grd_matrix} is treated in terms of two separate values which enables a convenient training procedure.

\subsection{Histograms}
We consider three different distance measures which are commonly used for histograms. In the following, the histograms for a central and a surrounding cell
w.r.t. channel $i$ are denoted as in Eq.~\ref{eq:eq_histogram_descriptor}. We compare the results of the three measurements in Section~\ref{sec:sec_experiments}.

\subsubsection{\textbf{Kullback-Leibler divergence}}
Using information theoretic arguments, one can represent the difference between a center and a surround cell using the Kullback-Leibler Divergence (KLD)~\cite{Kullback1951},
\begin{equation}
D_{KL}(P||Q)=\int_{-\infty}^{\infty}p(x)\ln\frac{p(x)}{q(x)}dx.
\end{equation}
Thus, the KLD between two probability distributions $P$ and $Q$  is a relative entropy that indicates the loss in information if $P$ is approximated by $Q$. The more $P$ differs from $Q$, the higher the KLD.

Given the histograms $H^i_c$ and $H^i_s$ of two channel vectors $P^i_c$ and $P^i_s$, we calculate the discrete KLD as our first contrast measure:
\begin{equation}
D_{KL}(H^i_c||H^i_s)= \sum\limits_{k=1}^{b} \ln\bigg(\frac{h_c^i(k)}{h_s^i(k)}\bigg) h_c^i(k).
\label{eq:eq_KLD}
\end{equation}

\subsubsection{\textbf{Hellinger distance}} Let $P$ and $Q$ be two probability distributions with respect to a probability measure $\lambda$; the Hellinger distance is a measure of their difference that is independent of $\lambda$. The square of the Hellinger distance has a particularly simple form and is defined as \cite{Hellinger1909}:
\begin{equation}
H^2(P,Q)=\frac{1}{2}\int\bigg(\sqrt{\frac{dP}{d\lambda}}-\sqrt{\frac{dQ}{d\lambda}} \bigg)^2d\lambda.
\end{equation}


For two discrete probability distributions $H^i_c$ and $H^i_s$ that represent $P^i_c$ and $P^i_s$, the Hellinger distance is then computed as the contrast between $P^i_c$ and $P^i_s$:
\begin{equation}
H^2(H^i_c,H^i_s)= \frac{1}{\sqrt{2}}\sqrt{\sum\limits_{k=1}^{b}\bigg(\sqrt{h_c^i(k)}-\sqrt{h_s^i(k)}\bigg)^2}.
\label{eq:eq_Hellinger}
\end{equation}

\subsubsection{\textbf{Histogram intersection}}
The histogram intersection is another popular similarity measure for histograms. Given two histograms $H_p$ and $H_q$ with $n$ bins, it is defined as:
\begin{equation}
\operatorname{HI}(H_{p},H_{q}) = \frac{\sum\limits_{k=1}^{n}\min(H_p(k), H_q(k))}{\sum\limits_{k=1}^{n}H_p(k)}.
\end{equation}

As all histograms considered in this paper are normalized so that they sum up to 1, the histogram intersection between $H^i_c$ and $H^i_s$ can be further simplified to:
\begin{equation}
\operatorname{HI}(H^i_c,H^i_s) =
\sum\limits_{k=1}^{b}\min(h_c^i(k), h_s^i(k)).
\label{eq:eq_hist_intersection}
\end{equation}

\section{Classification}
\label{sec:sec_classification}
In this section, we discuss our approach towards classification of the center-surround features introduced above. First of all, we address the size of our feature pool. Given a pedestrian model of $60\times 120$ pixels, Tab.~\ref{tab:tab_feature_size} compares feature sizes under different settings in terms of scales and dimensions of contrast vectors. Apparently, the feature pool grows once more scales are employed. Among all the contrast measurements considered in this paper, only the signed gradient matrix is two dimensional, while all others are one dimensional.

\begin{table}\small
\renewcommand{\arraystretch}{1.5}
\centering
\begin{tabular}{ l ||  c | c | c }
\toprule[1.5pt]
Scales & 4-6 & 4-6-8 & 4-6-8-10 \\
\hline
Feature size  & $20,320D(\overrightarrow{cst})$ & $23,440D(\overrightarrow{cst})$ &$25,040D(\overrightarrow{cst})$\\
\bottomrule[1.5pt]
\end{tabular}
\caption{Illustration of feature size under different configurations. All the contrast measurements used in this paper are one dimensional, except SGrd, which is two dimensional.}
\label{tab:tab_feature_size}
\end{table}

To efficiently train classifiers on such a large feature pool, we
employ a fast version of AdaBoost \cite{Appel2013} since it offers a convenient and fast approach to feature selection from a large number of candidate features.
The feature selection procedure is
conducted for each feature configuration individually.

For boosting algorithms, one should choose proper weak classifiers so as to build the final strong classifier. We use decision trees of depth 2 as our weak classifiers since they are efficient to learn. Another important parameter is the number of weak classifiers, which, after extensive experimentation, we choose to be 4096, as we observe that more weak classifiers do not lead to gains in performance. Similar to classic approaches to pedestrian detection~\cite{Dalal2005,Dollar2009a}, we also employ a multi-round training strategy which has been shown to lead to better performance than a simple one round training procedure with the same number of
samples. For the first round, the initial negative training samples are randomly cropped from the negative example images; in the following rounds, hard negative samples are exhaustively searched over all negative example images using the classifier built in the previous round. This procedure is iterated until no significant performance gains are observed with further retraining. From our experiments, three rounds of retraining
yield optimal performance; additional rounds did not show significant improvements. We collect 5000 negative samples at each round, resulting in a large negative sample pool of 20,000 image patches after four rounds.

\begin{figure}[t]
\centering
\begin{subfigure}[t]{0.18\textwidth}
\centering
	\includegraphics[width=\textwidth]{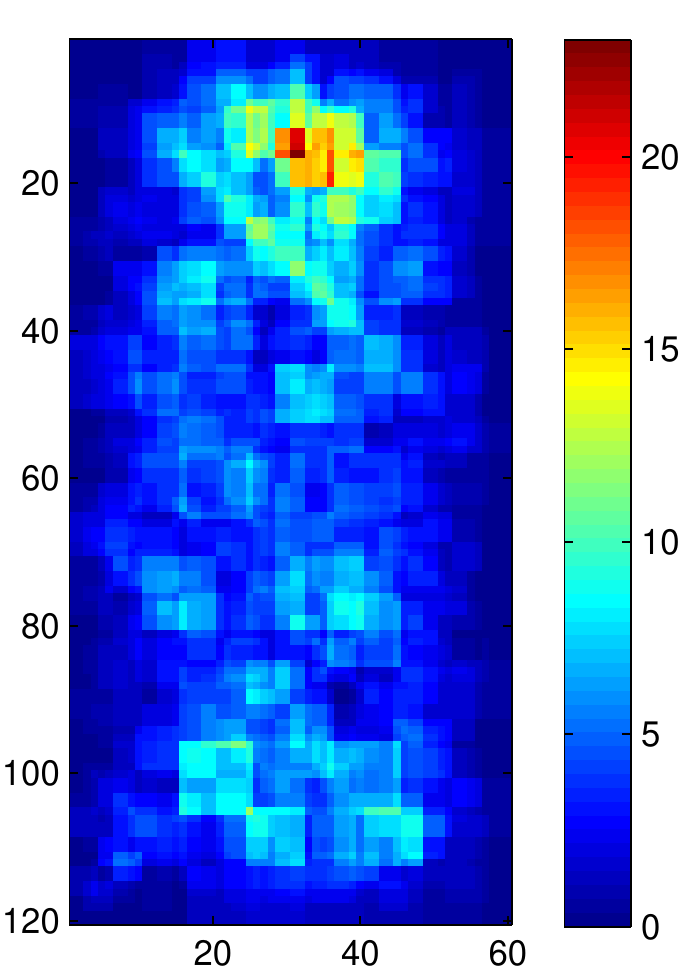}
	\caption{}
	\label{fig:fig_cell_weight}
\end{subfigure}
\hspace{1pt}
	\begin{subfigure}[t]{0.18\textwidth}
\centering
	\includegraphics[width=1.09\textwidth]{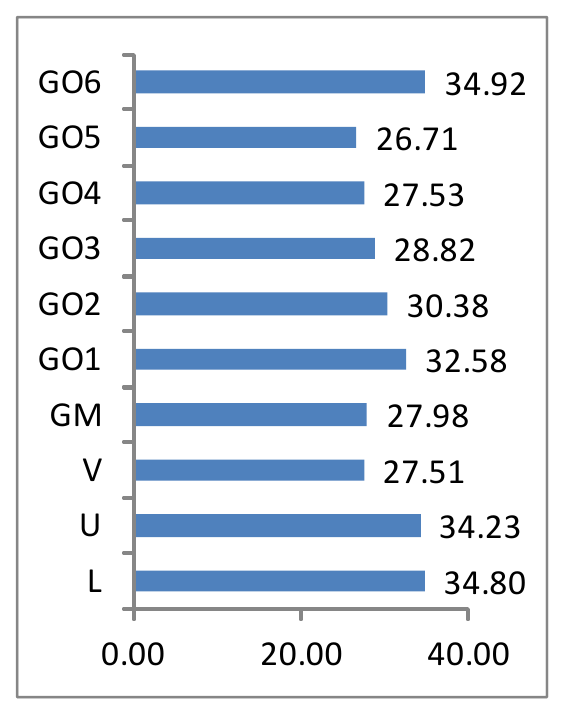}
	\caption{}
	\label{fig:fig_channel_weight}
\end{subfigure}
\caption{Illustration of representative features under one configuration. (a) Body parts weight map: different colors are used to indicate the accumulative weight of each pixel after boosting. (b) Channel weight bars: accumulative weight of each channel is indicated by one bar.}
\label{fig:fig_representative_features}
\end{figure}

In order to look into which local features regions are more informative, we plot a weight image of the top 100 feature positions with highest weights from the final strong classifier, as shown in Fig.~\ref{fig:fig_cell_weight}. To generate this map, we add the weight of each selected feature to the cells it covers and use different colors to indicate the accumulative weight of each cell after boosting. As expected, the head-shoulder area of the human body shows to be more discriminative for pedestrian detection than other body parts. Moreover, we also add the weight of each feature separated by channels to indicate which ones are more representative and use bars to illustrate the accumulative weight of each channel as shown in Fig.~\ref{fig:fig_channel_weight}. We find that all the channels we chose contributed rather evenly to the final classifier, indicating no channel redundancy.

The most discriminative features selected by the boosting algorithm are then used for pedestrian detection in still images. Our pedestrian model is of size $60\times 120$ pixels and we resize the input image to detect pedestrians at different scales. To this end, we slide a window over the whole image and consider multiple scales. The spatial step size is set identical to the cell size for speed reasons and the scale step is set to be $1.09$ so that there are $8$ scales per octave. We use a simplified non-maximal suppression (NMS) procedure \cite{Dollar2009a} to suppress nearby detections.

\section{Experiments}
\label{sec:sec_experiments}
In this section, we introduce the benchmark datasets and evaluation protocols used in our experiments, provide comprehensive comparisons for different feature schemes, and compare our
best detector configuration with state-of-the-art detectors.

\subsection{Benchmark datasets}
Experiments are conducted on two public benchmark datasets: the INRIA Person Dataset~\cite{Dalal2005} and the Caltech Pedestrian Detection Benchmark~\cite{Dollar2011}. A comparison of the above two datasets is given in Tab.~\ref{tab:tab_datasets}.

\textit{INRIA Person Dataset}: This is arguably the most popular dataset for people detection and comes along with pre-defined subsets for training and testing. In the training set, there are 2416 positive samples, by mirroring from 1208 identical pedestrian images; and 12,180 natural images, where no pedestrians are included so that negative samples can be randomly generated by cropping subregions. In the test set, there are 288 positive samples, including 566 pedestrian annotations.

\textit{Caltech Pedestrian Detection Benchmark}: This is currently the largest and most challenging dataset for pedestrian detection, consisting of approximately 10 hours of 640$\times$480 30Hz video taken from a vehicle driving through regular traffic in an urban environment. About 250,000 frames with a total of 350,000 bounding boxes and 2300 unique pedestrians were annotated. The training data (set00-set05) and the test data (set06-set10) consist of approximately $192,000$ and $155,000$ pedestrian annotations, respectively.

\begin{table} \small 
\centering
\begin{tabular}{ c || c | c |c }
\toprule[1.5pt]
& & INRIA \cite{Dalal2005} & Caltech \cite{Dollar2009b} \\
\bottomrule[1.5pt]
\multirow{4}{*}{Properties} & imaging setup & photo & mobile\\
& {color images} & $\surd$ & $\surd$ \\
& {video seqs.} & $\times$ & $\surd$ \\
& {occlusion labels} & $\times$ & $\surd$ \\
\hline
\multirow{3}{*}{Training} & \# pedestrians & 1208 & 192k\\
& \# pos. images & 614 & 67k\\
& \# neg. images & 1218 & 61k\\
\hline
\multirow{3}{*}{Testing}& \# pedestrians & 566 & 155k\\
& \# pos. images & 288 & 65k\\
& \# neg. images & 453 & 56k\\
\bottomrule[1.5pt]
\end{tabular}
\caption{Statistics of two pedestrian datasets used for experiments~\cite{Dollar2011}.}
\label{tab:tab_datasets}
\end{table}

\subsection{Evaluation protocol}
\label{sec:sec_evaluation_protocol}
In the following, we explain details of our evaluation protocol in four
aspects, which are consistent with the conventions in this field~\cite{Dollar2011}.

\subsubsection{Ground truth filtering} In our experiments, a \textit{reasonable} subset of all ground truth data is considered, in which pedestrians at a resolution of over 50 pixels in height and a visibility of more than 65\% are considered. Outliers are marked with an \textit{ignore} label, which means they need not be matched, however, matches are not considered as mistakes either.

\subsubsection{Detection results filtering} We filter out detection results using an expanded filtering method~\cite{Dollar2011}, so that detection results far outside the evaluation scale range should not be considered. In this paper, we evaluate a scale range of $[50, +\infty]$, only detections in $[50/\xi,+\infty]$ are considered for evaluation. In our experiments, we set $\xi = 1.25$~\cite{Dollar2011}.

\subsubsection{Bounding box matching rules} A filtered ground truth bounding box and detection bounding box are annotated by $B_\text{gt}$, and $B_\text{dt}$ respectively. $B_\text{gt}$, and $B_\text{dt}$ match if and only if the ratio of overlap to the union of their areas exceeds a given threshold~\cite{Dollar2011}:
\begin{equation}
\operatorname{match}(B_\text{dt},B_\text{gt}) = \frac{\operatorname{area}(B_\text{dt})\cap\ \operatorname{area}(B_\text{gt})}{\operatorname{area}(B_\text{dt})\cup\ \operatorname{area}(B_\text{gt})}\stackrel{!}{>} 0.5 \text{ .}
\label{eq:overlap}
\end{equation}

\subsubsection{Performance measurements} We perform evaluation w.r.t. full images instead of detection windows as the former one provides a natural measure of error of an overall detection system. In this paper, we employ two measurements to compare performance among different detectors. First, we plot miss rate against false positives per image (FPPI) curves in logarithmic scales by varying the threshold on the detection confidence of the classifiers. In addition to this miss rate vs. FPPI curves, we calculate a single, numerical measurement to summarize each detector's performance. We use the \textit{average miss rate}~\cite{Dollar2011}, which is computed by averaging the miss rates at nine FPPI rates evenly sampled in log-space in the range of $[10^{-2}, 10^0]$. This \textit{average miss rate} generally gives a more stable and informative assessment of the overall performance for different detectors than the miss rate at only $10^{-1}$ FPPI~\cite{Dollar2011}.

\subsection{Comparisons for different feature settings}
\label{sec:sec_comparisons_schemes}
In this section, we seek the strongest feature scheme through experiments under different settings on the INRIA dataset. First, we define a default setting with: three scales of $4\times 4$, $6\times 6$, and $8\times 8$ pixels; 5 histogram bins when histograms are used.

In the following, we compare different descriptors, contrast measurements, scale structures, and numbers of histogram bins where histograms are used.

\subsubsection{Contrast measurements} 
\begin{figure}[t]
\centering
\begin{subfigure}[t]{0.4\textwidth}
	\centering
	\includegraphics[width=\textwidth]{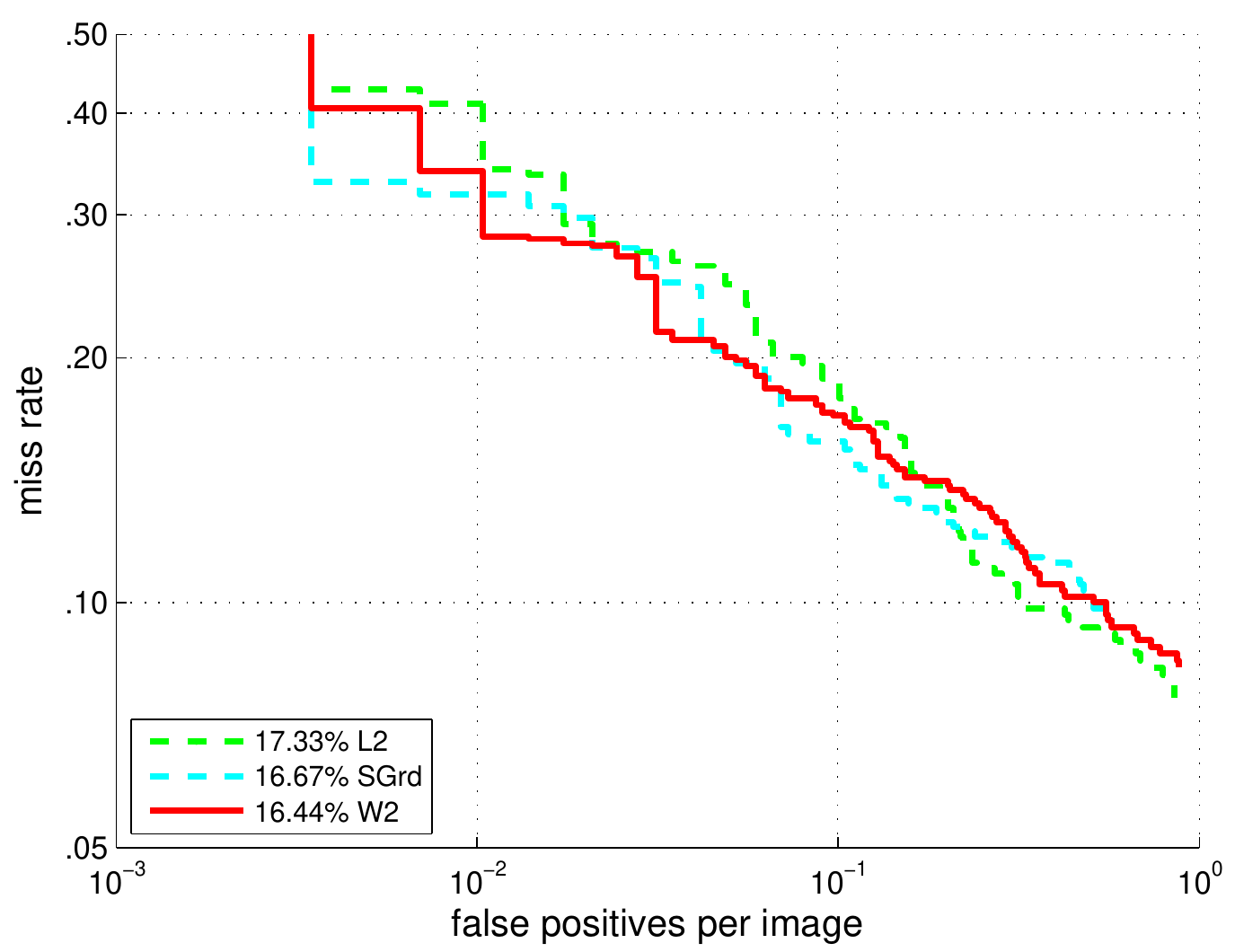}
	\caption{Gaussian distributions}
	\label{fig:fig_gaussians}
\end{subfigure}
\hfill
\begin{subfigure}[t]{0.4\textwidth}
	\centering
	\includegraphics[width=\textwidth]{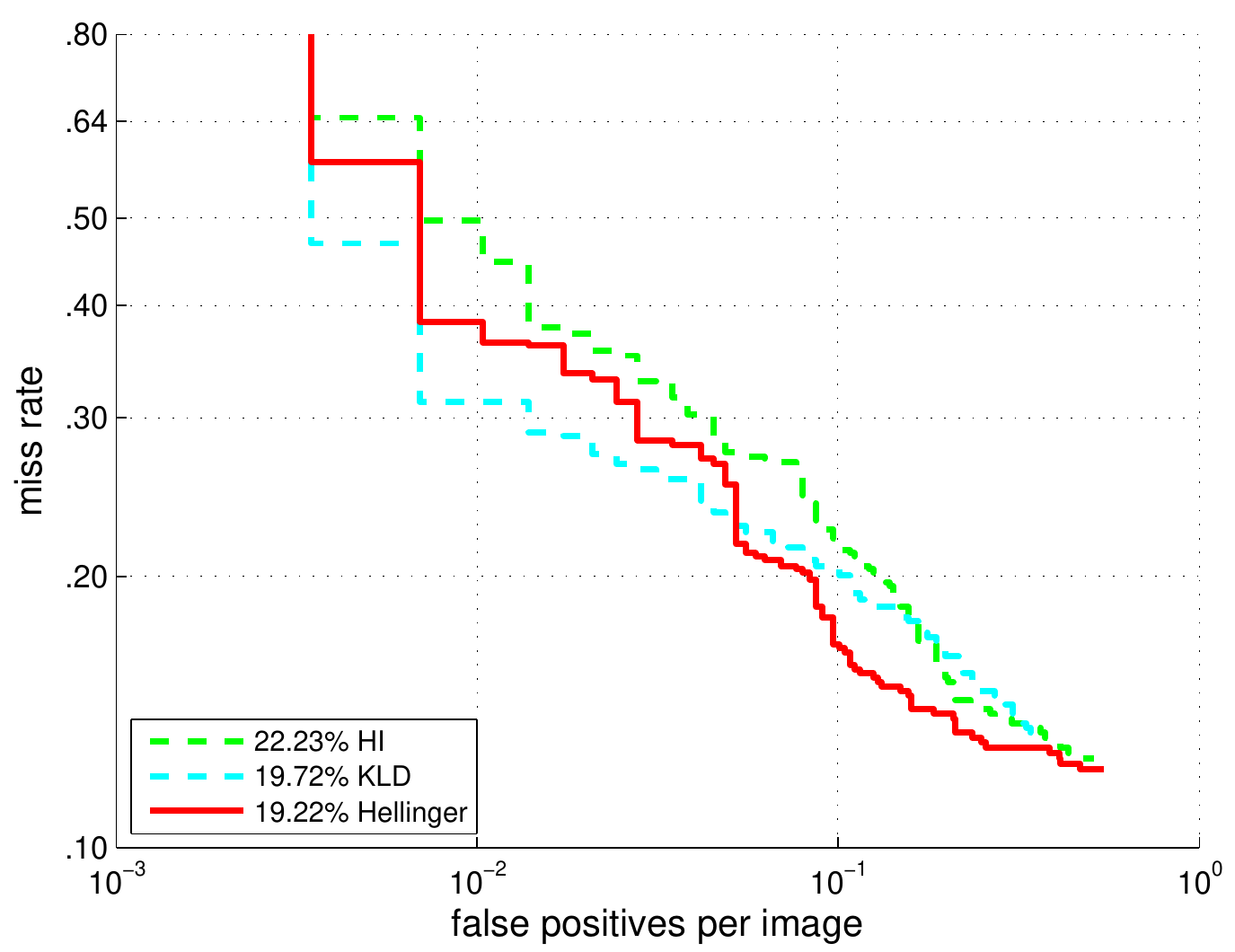}
	\caption{Histograms}
	\label{fig:fig_hists}
\end{subfigure}
\caption{Experiments on different contrast measures for two cell descriptors.}
\label{fig:fig_measurements}
\end{figure}

We investigate different contrast measures for two descriptors respectively. From Fig.~\ref{fig:fig_gaussians}, we see that both descriptors produce stable results using different contrast measures. Despite of their stable performance, we observe a slight difference between different contrast measures. For Gaussian descriptors, $L^2$ distance performs worst, and \wdist~distance and SGrd produce comparably better results. For histograms, HI is the worst measure, and KLD and Hellinger are comparably better. Therefore, we select Gaussian-\wdist~and Hist-Hellinger as the two preferable combinations which produce best results for Gaussian and Histogram descriptors, respectively.

\subsubsection{Number of histogram bins} 
\begin{figure}
	\centering
	\centerline{\includegraphics[width=0.4\textwidth]{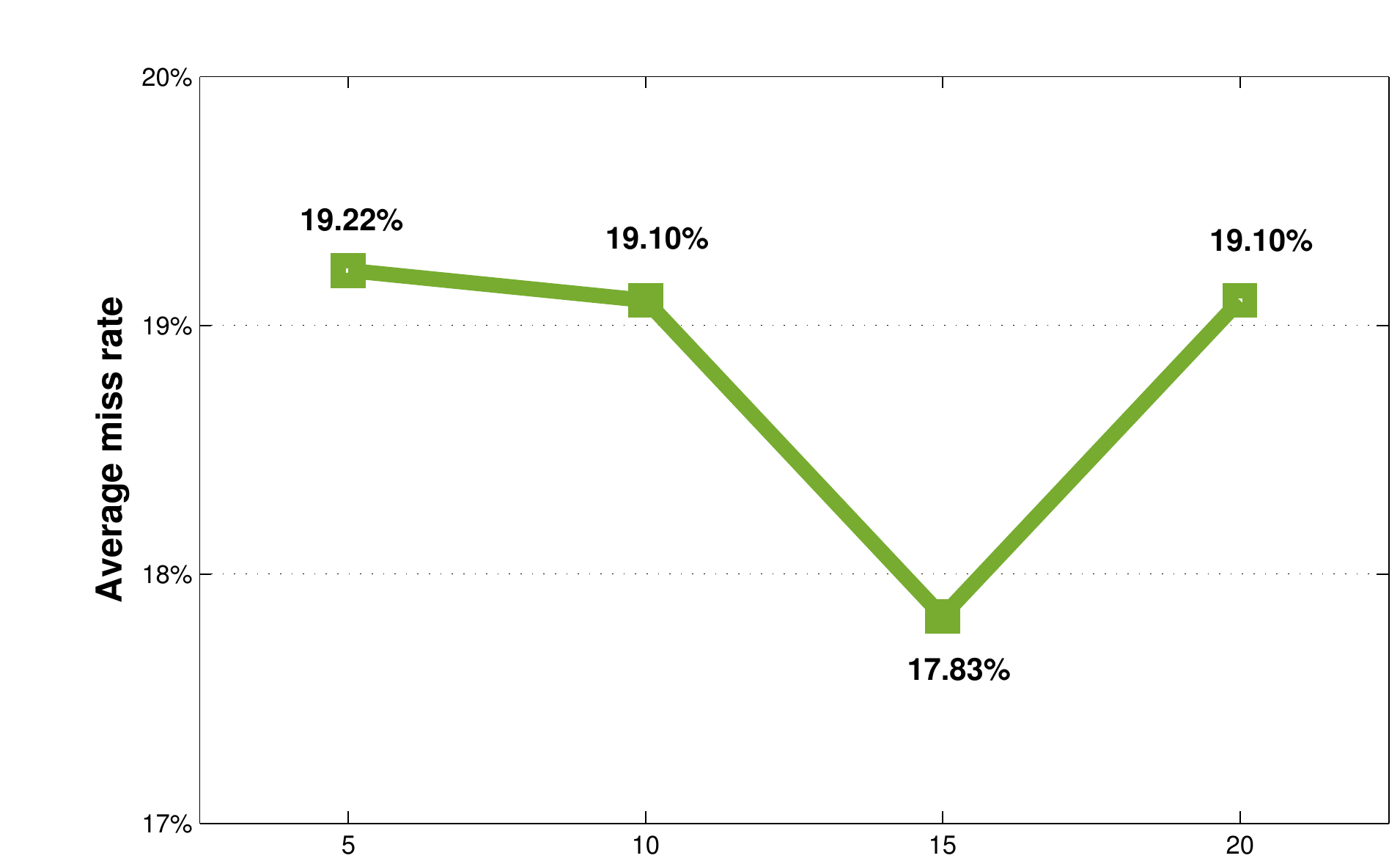}}
	\caption{Experiments with different histogram bins using the Hellinger distance.}
	\label{fig:fig_histogram_bins}
\end{figure}

The number of bins is an important parameter in practical applications of histograms. Not surprisingly, we observe performance changes when increasing the number of histogram bins. Fig.~\ref{fig:fig_histogram_bins} shows experimental results when using 5, 10, 15 and 20 bins and the Hellinger distance which has been shown to be the best among all the contrast measures for histograms. Generally, more histogram bins integrate more accurate information of the local cell region, thus leading to better performance. If we increase the number of histogram bins from 5 to 15, as expected we obtain better results. However, performance begins to decrease again when we consider more than 20 bins since these settings are more error prone under noisy real world data. Note that, in the following experiments, we thus use 15-bin histograms instead of the default 5-bin histograms.

\subsubsection{Descriptors} 
\begin{figure}
	\centering
	\centerline{\includegraphics[width=0.4\textwidth]{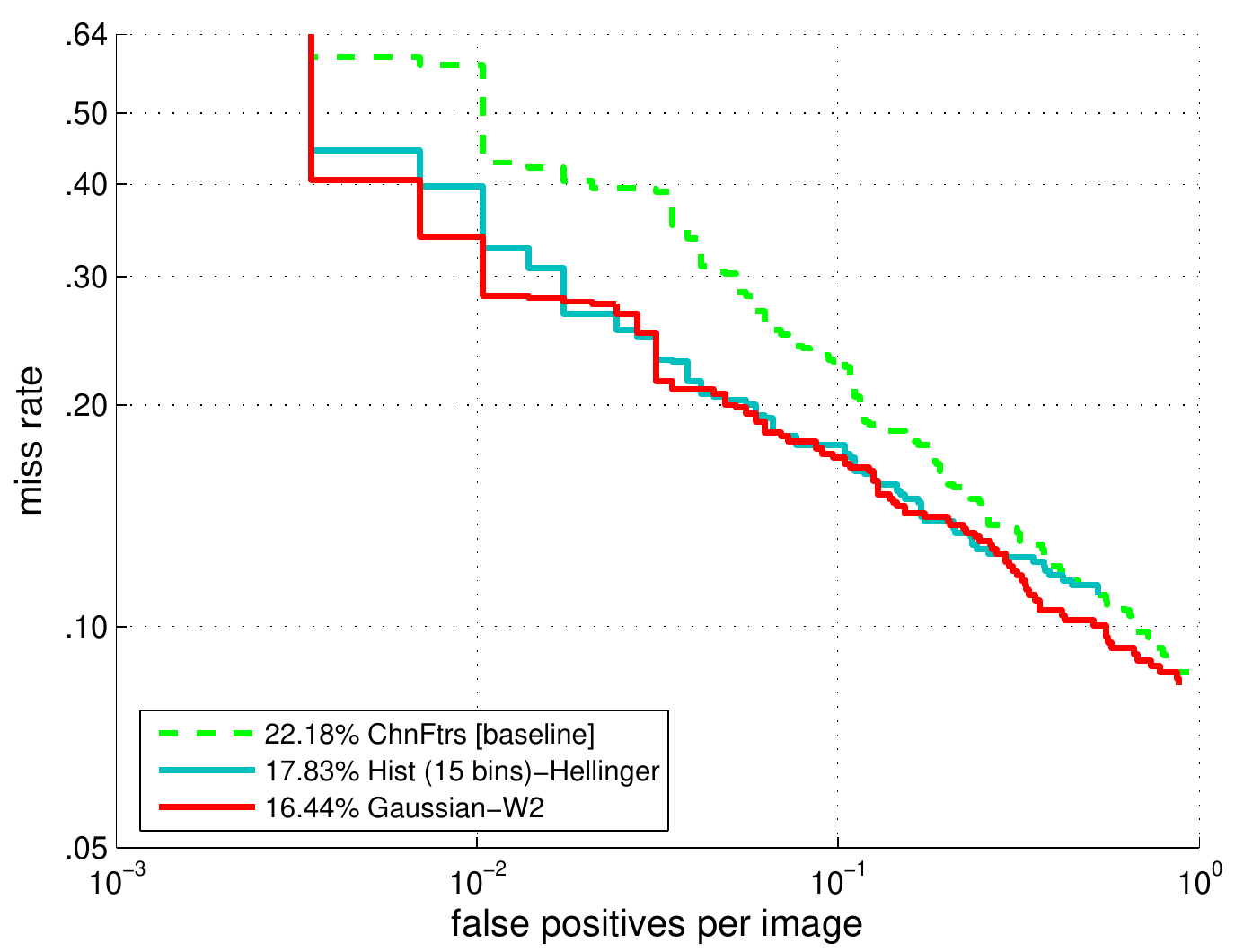}}
	\caption{Comparison of two optimal descriptor-measurement combinations and the baseline detector [ChnFtrs].}
	\label{fig:fig_descriptors}
\end{figure}

From Fig.~\ref{fig:fig_descriptors}, we can see that both optimal combinations outperform the baseline detector [ChnFtrs] which illustrates the effectiveness of our new features. Between the two optimal combinations, Gaussian-\wdist produces better overall results than Hist(15 bins)-Hellinger. Therefore, Gaussian-\wdist~is selected as the optimal descriptor-contrast-measurement combination in this paper.

\subsubsection{$C_1S_8$ pattern vs. $C_1S_1$ pattern} 
\begin{figure}
	\centering
	\centerline{\includegraphics[width=0.5\textwidth]{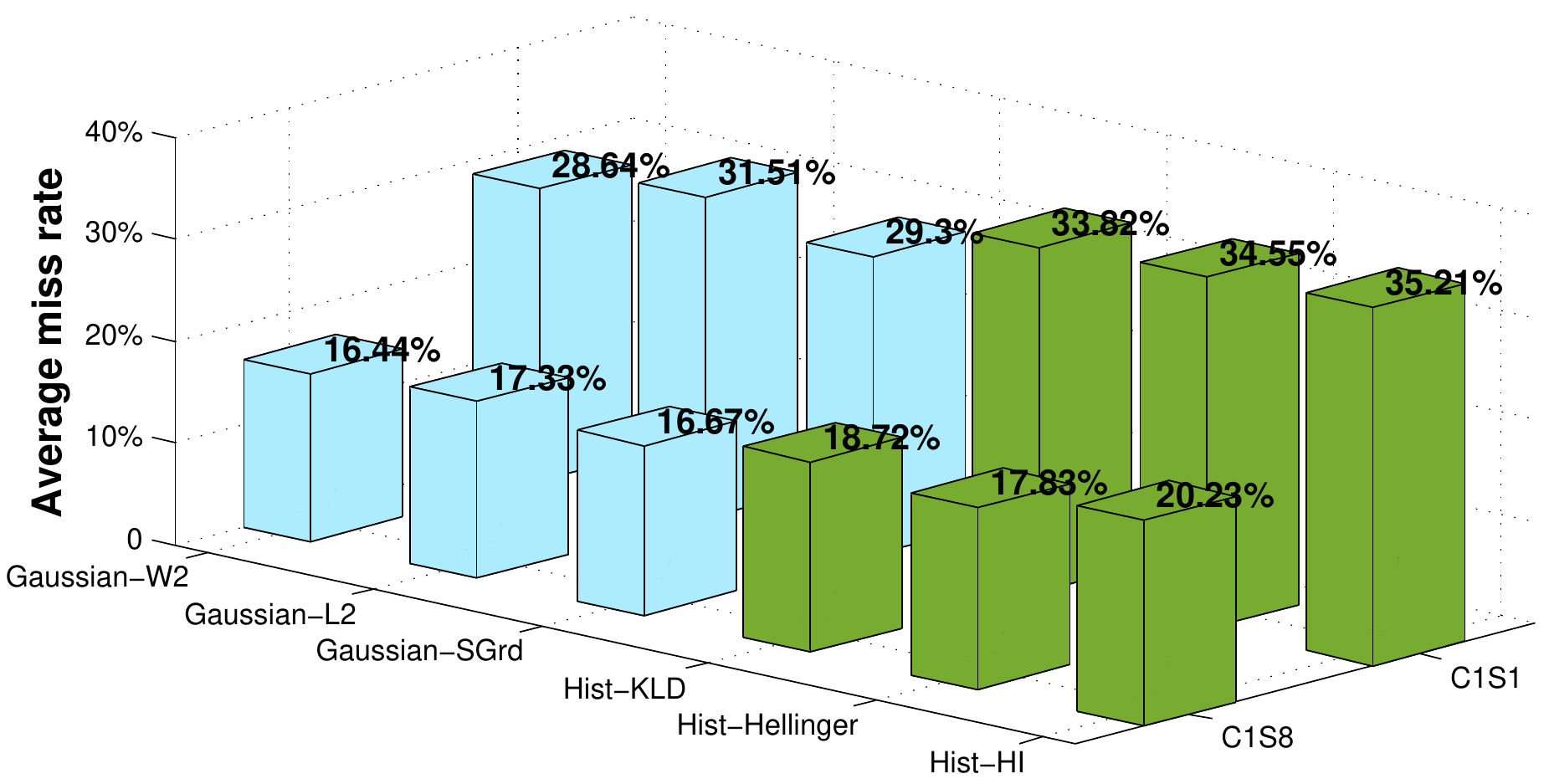}}
	\caption{Comparison of two center-surround patterns.}
	\label{fig:fig_cspatterns}
\end{figure}

We proposed the $C_1S_8$ pattern in Section~\ref{sec:sec_overview} in order to incorporate more information about local image differences. Here, we compare the performance of both patterns to show why the directed $C_1S_8$ pattern is superior. From Fig.~\ref{fig:fig_cspatterns}, it appears that the $C_1S_8$ pattern produces better results than $C_1S_1$ over all descriptor-contrast-measurement combinations.

\subsubsection{Scale structures} 
\begin{figure}
	\centering
	\centerline{\includegraphics[width=0.5\textwidth]{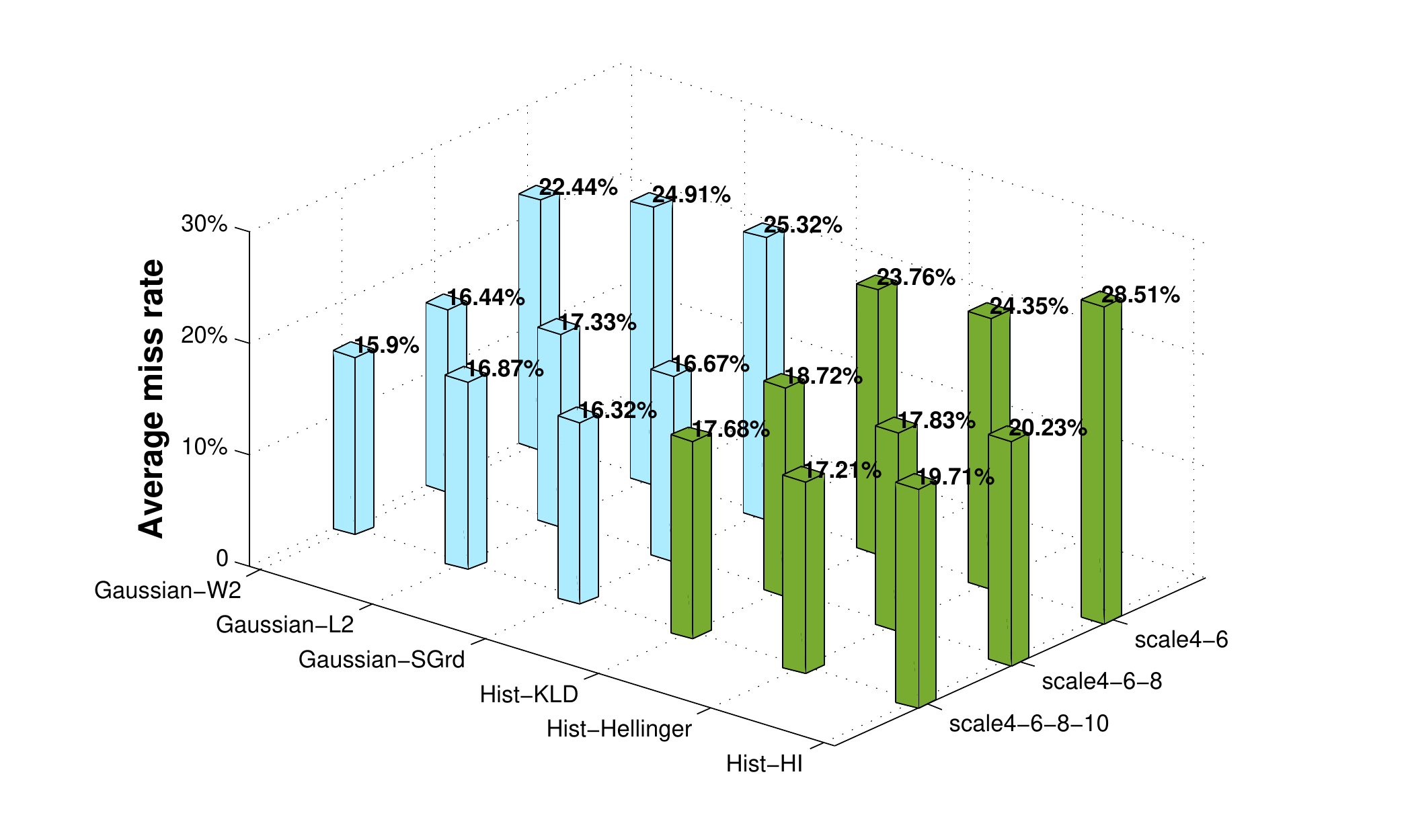}}
	\caption{Comparison of three scale structures. In this experiment, 15-bin histograms are used.}
	\label{fig:fig_scale_structures}
\end{figure}

Generally, the use of more scales incorporates richer information and leads to a better performance. In this paper, we consider three different scale structures: 4-6; 4-6-8; and 4-6-8-10, and show their comparisons in Fig.~\ref{fig:fig_scale_structures}. Increasing the scales from 4-6 to 4-6-8 brings about a significant improvement of approximately $5\%$ w.r.t. miss rate; on the other hand, continuing to increase scales to 4-6-8-10 produces a less prominent performance gain of less than $1\%$ w.r.t. miss rates. Therefore, we choose a scale structure of 4-6-8-10 as our optimal choice.

In summary, the optimal feature setting is to use the combination of Gaussian-\wdist~and scale structure of 4-6-8-10. We use this configuration in the following experiments.

\subsection{Computational complexity} We investigate the computational complexity of different feature settings.
Our normal-distribution as well as histogram based features are computed from local averages of certain values. Such local features can be computed in $\mathcal{O}(n)$ time with $n$ denoting the number of image pixels using \textit{moving averages} or \textit{integral image} techniques.  They only differ in the number of layers needed (one for each distribution parameter or bin) which amounts to a constant factor.
Looking into details of the diverse distance functions implemented for different feature settings, we can see the very same effect: the time complexity is constant per pixel (linear growing with image size), so the overall complexity for each setting is still $\mathcal{O}(n)$. We have to note that the constant factor for normal-distributions is $2$ per input channel, while histograms require $b\geq2$ (\eg $15$) number of histogram bins.

The computational complexity of our baseline detector [ChnFtrs] is also $\mathcal{O}(n)$, because each pixel is visited once per channel for computing local sums. Therefore, the computational complexity of our features at different settings is on par with [ChnFtrs].


\subsection{Comparisons with state-of-the-art detectors}
\begin{table}[t] \small 
\setlength{\tabcolsep}{15pt}
\renewcommand{\arraystretch}{1.2}
\centering
\begin{tabular}{ l || l| l }
\toprule[1.5pt]
Detector & \multicolumn{2}{c}{Average miss rate}\\
    &    INRIA & Caltech\\
\toprule[1.5pt]
VJ\cite{Viola2004} & 72.48\% & 94.73\%\\
\hline
HOG\cite{Dalal2005} & 45.98\% & 68.46\%\\
\hline
Shapelet\cite{Sabzmeydani2007} & 81.70\% & 91.37\%\\
\hline
MultiFtr~\cite{Wojek2008} & 36.50\% & 68.62\%\\
\hline

MultiFtr+CSS~\cite{Walk2010} & 24.74\% & 60.89\%\\
\hline
MultiFtr+Motion~\cite{Walk2010} & \quad / & 50.88\%\\
\hline
HikSvm~\cite{Maji2008} & 42.82\% & 73.39\%\\
\hline
HogLbp~\cite{Wang2009} & 39.10\% & 67.77\%\\
\hline
LatSvm-V1~\cite{Felzenszwalb2008} & 43.83\% & 79.78\%\\
\hline
LatSvm-V2~\cite{Felzenszwalb2010} & 19.96\% & 63.26\%\\
\hline
ChnFtrs~\cite{Dollar2009a} & 22.18\% & 56.34\%\\
\hline
FeatSynth~\cite{Bar-Hillel2010} & 30.88\% & 60.16\%\\
\hline
MultiResC~\cite{Park2010} & \quad / & 48.45\%\\
\hline
CrossTalk~\cite{Dollar2012} & 18.98\% & 53.88\%\\
\hline
VeryFast~\cite{Benenson2012} &  15.96\% & \quad /\\
\hline
SketchTokens~\cite{Lim2013} & 13.32\%$\bigstar$ & \quad /\\
\hline
Roerei~\cite{Benenson2013} & 13.53\%$\bigstar$ & 48.35\%\\
\hline
RandForest~\cite{Marin2013} & 15.37\%$\bigstar$ & 51.17\%\\
\hline
AFS+Geo~\cite{Levi2013} & \quad / & 66.76\%\\
\hline
MT-DPM+Context~\cite{Yan2013}& \quad / & 37.64\%$\bigstar$\\
\hline
DBN-Isol~\cite{Ouyang2012} & \quad / &53.14\%\\
\hline
DBN-Mut~\cite{Ouyang2013a} & \quad / & 48.22\%\\
\hline
ACF+SDt~\cite{Park2013} & \quad / & 37.34\%$\bigstar$\\
\bottomrule[1.5pt]
ours & 15.90\% & 34.96\%$\bigstar$\\
\bottomrule[1.5pt]
\end{tabular}
\caption{Performance comparisons to state-of-the-art pedestrian detectors. Each row in this table displays the corresponding average performance in terms of average miss rates. The approach proposed in this paper yields state-of-the-art performance on the INRIA dataset and consistently better results than previously reported methods on the Caltech dataset. We indicate the top three detectors for each dataset with $\bigstar$.}
\label{tab:tab_detectors}
\end{table}

In this section, we compare the performance of our detector with optimal settings
found in Section~\ref{sec:sec_comparisons_schemes} to state-of-the-art detectors whose results are publicly available\footnote{\burl{http://www.vision.caltech.edu/Image_Datasets/CaltechPedestrians/}}
using the experimental protocol explained in Section~\ref{sec:sec_evaluation_protocol}.

The results on the INRIA dataset in Fig.~\ref{fig:fig_results_inria} show that our detector outperforms the baseline detector [ChnFtrs] by about 6\% and reaches state-of-the-art performance. On the Caltech pedestrian dataset, our detector outperforms not only the baseline detector [ChnFtrs] by about 15\% but also yields the overall best performance as shown in Fig.~\ref{fig:fig_results_caltech}. More extensive comparisons are shown in Tab.~\ref{tab:tab_detectors}.

\begin{figure*}[t]
\begin{subfigure}[b]{0.3\textwidth}
	\centering
	\includegraphics[width=\textwidth]{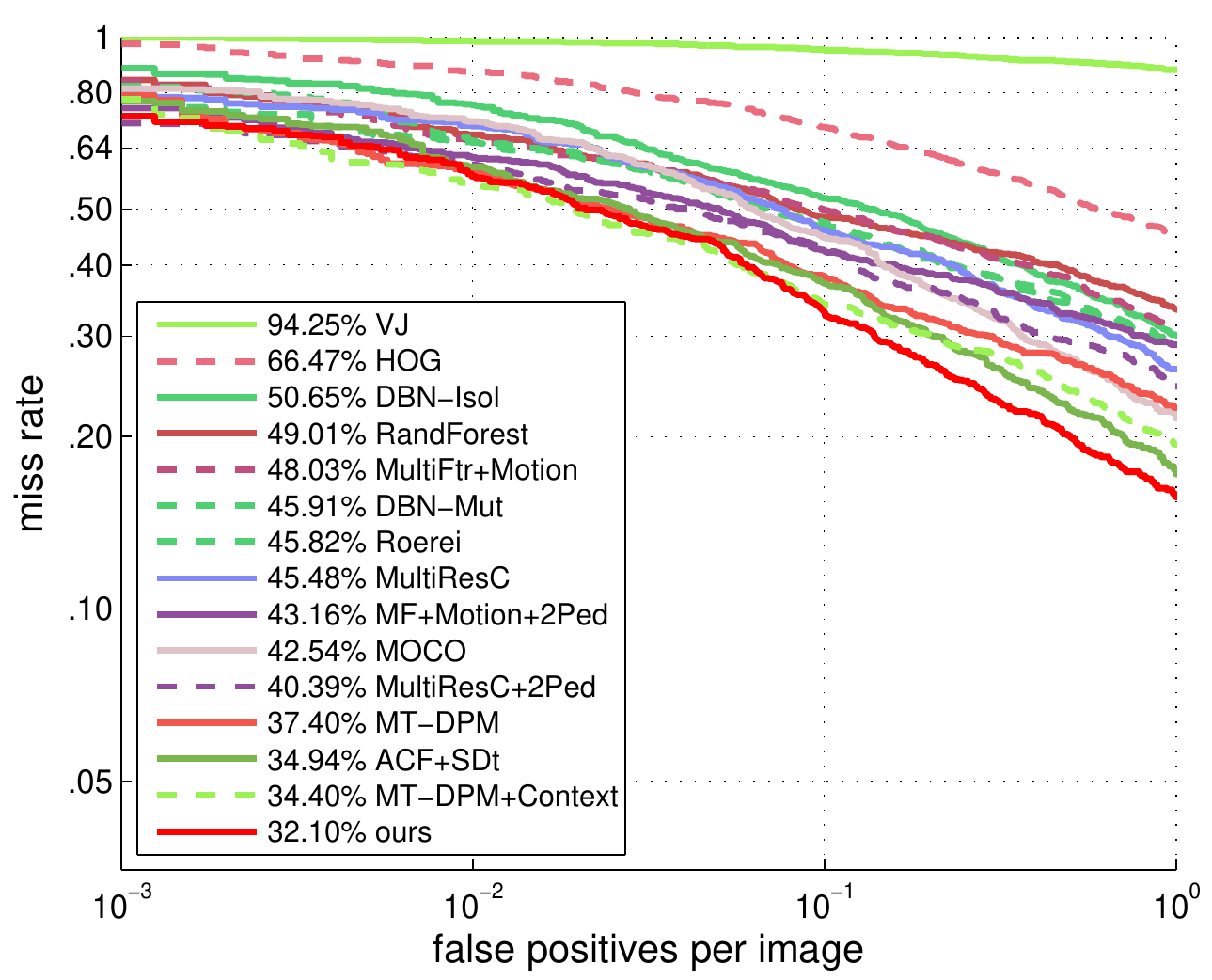}
	\caption{No occlusion (100\% visible)}
	\label{fig:fig_results_inria}
\end{subfigure}
\hfill
\begin{subfigure}[b]{0.315\textwidth}
	\centering
	\includegraphics[width=\textwidth]{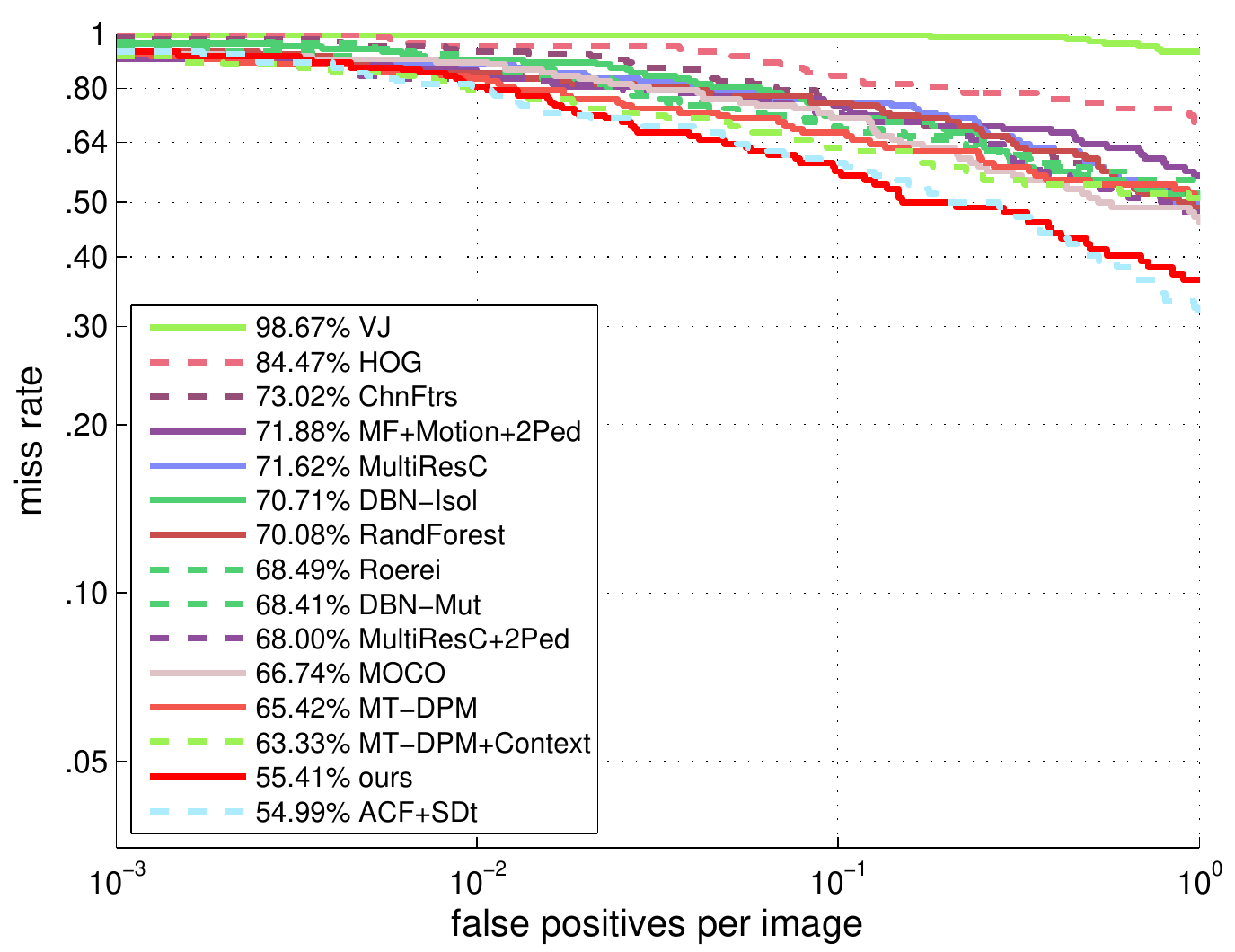}
	\caption{Partial occlusion (65\%-100\% visible)}
	\label{fig:fig_results_caltech}
\end{subfigure}
\hfill
\begin{subfigure}[b]{0.3\textwidth}
	\centering
	\includegraphics[width=\textwidth]{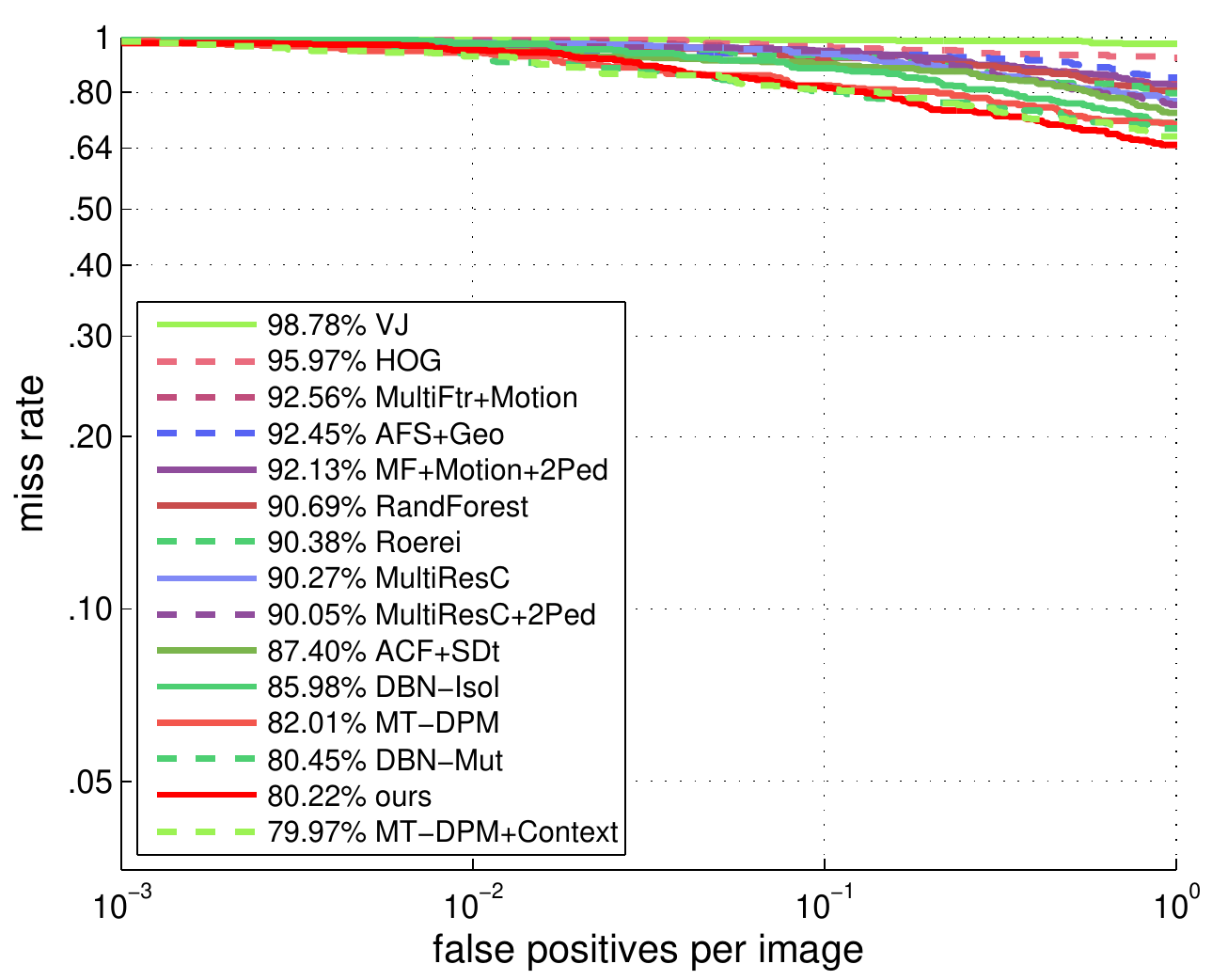}
	\caption{Heavy occlusion (20\%-65\% visible) }
	\label{fig:fig_results_caltech}
\end{subfigure}
\caption{Results on Caltech dataset under different occlusion conditions.}
\label{fig:fig_results_occlusion}
\end{figure*}

Moreover, we show results under different occlusion conditions on the Caltech dataset, as shown in Fig.~\ref{fig:fig_results_occlusion}. Our detector ranks stably high across different occlusion levels, and outperforms some part-based approaches, \eg MT-DPM. These evaluations demonstrate that our approach is robust to occlusions. The major reason is that most informative features are automatically selected from the head-shoulder area (as shown in Fig.~\ref{fig:fig_cell_weight}), where occlusion is less likely to happen.

\begin{figure*}[t]
\begin{subfigure}[b]{0.5\textwidth}
	\centering
	\includegraphics[width=1.0\textwidth]{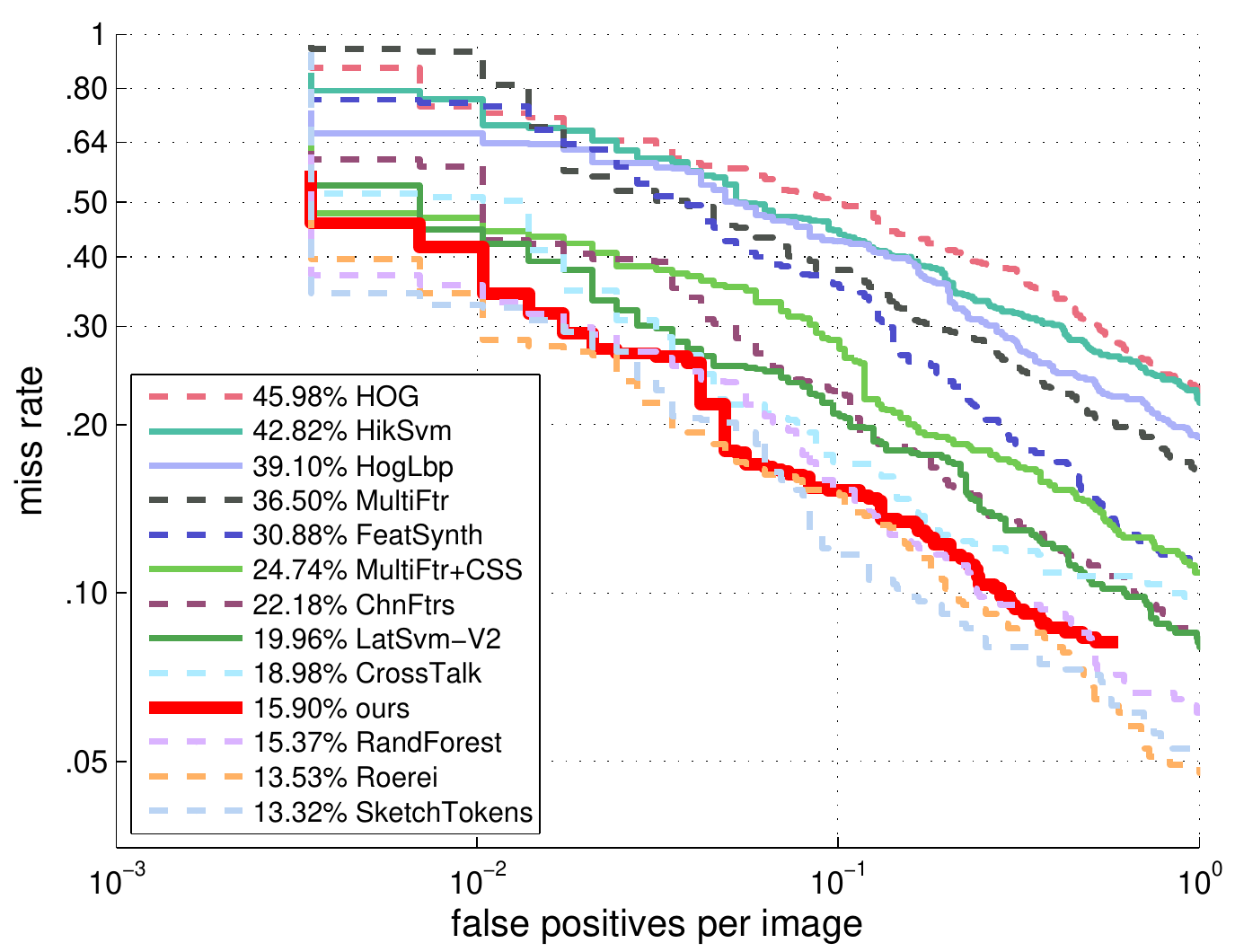}
	\caption{INRIA}
	\label{fig:fig_results_inria}
\end{subfigure}
\hfill
\begin{subfigure}[b]{0.5\textwidth}
	\centering
	\includegraphics[width=\textwidth]{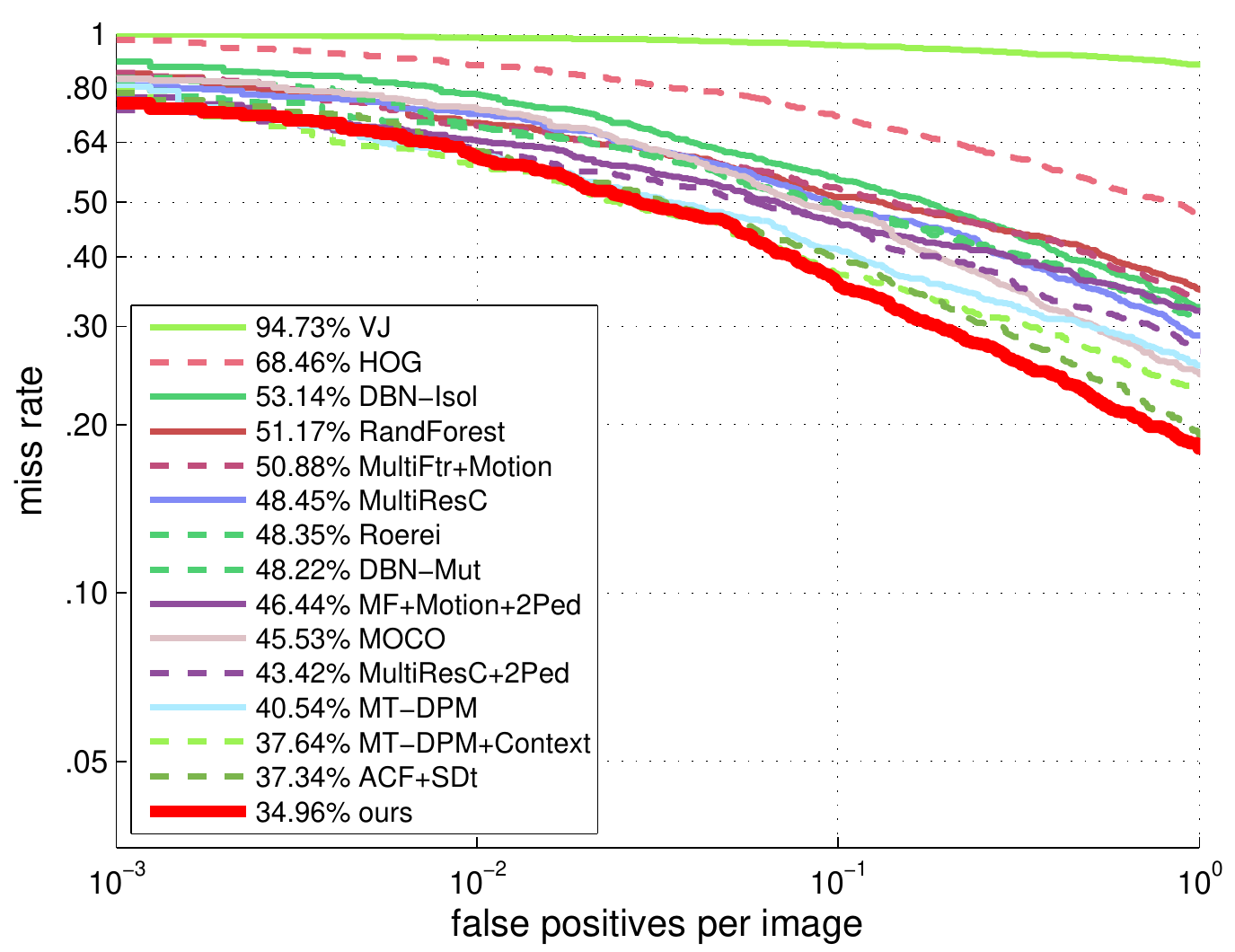}
	\caption{Caltech}
	\label{fig:fig_results_caltech}
\end{subfigure}
\caption{Overall results of different detectors on the INRIA and Caltech datasets under standard evaluation settings.}
\label{fig:fig_results_alldetectors}
\end{figure*}

\section{Conclusion}
\label{sec:sec_conclusion}
Humans are able to efficiently locate what they are looking for because the human visual system is tuned to characteristic visual features so that objects of interest become salient. This mechanism is called top-down saliency or visual search. In this paper, we tried to mimic early human visual processing by using local distribution contrast features and boosted them to respond to the appearance of pedestrians. The resulting pedestrian detector thus realizes a computational top-down saliency system.
Our features are very efficient to compute by means of combining a fast integral method for local averaging and a clever arrangement of additional image layers for fast maximum likelihood estimation of parameters of normal distributions. We tested different patterns for organizing the center-surround structure and scale structure as well as different ways to estimate the cell distribution and contrast measurements.


Experimental results showed that our detector achieves state-of-the-art performance on the INRIA pedestrian dataset. Moreover, on the Caltech pedestrian dataset, we found it to outperform all other recent approaches considered.

Given these results, it appears promising to further explore feature design driven by human visual mechanisms. Immediate extensions of the approach presented in this paper consist in incorporating information from additional channels such as motion and depth. This is currently explored in ongoing work and results will be reported once they become available.


%

%
%

\ifCLASSOPTIONcaptionsoff
  \newpage
\fi



%
{
\bibliographystyle{IEEEtran}
\bibliography{literature}
}

%




\end{document}